\documentclass[10pt,final,twocolumn,twoside]{IEEEtran}

\usepackage[normalem]{ulem}
\usepackage{cite}
\usepackage{url}
\usepackage{graphicx}
\usepackage{booktabs}
\usepackage{tabularx}
\usepackage{multirow}
\usepackage{subfigure}
\usepackage{amsmath,dsfont}
\usepackage{array}
\usepackage{subfigure}
\usepackage{mathtools}

\usepackage{algorithm,algorithmic}

\usepackage{verbatim}
\usepackage{bbm}

\usepackage{colortbl}
\definecolor{mygray}{gray}{.75}

\usepackage{scalerel,stackengine}

\usepackage{framed}
\usepackage{color}
\usepackage{xcolor}
\usepackage{amssymb}

\newcommand{\red}[1]{\textcolor{red}{#1}}

\usepackage[normalem]{ulem}

\newcommand{\etal}{\emph{et al.}~}
\newcommand{\ie}{\emph{i.e.,}~}

\newcommand\openbigstar[1][0.7]{%
  \scalerel*{%
    \stackinset{c}{-.125pt}{c}{}{\scalebox{#1}{\color{white}{$\bigstar$}}}{%
      $\bigstar$}%
  }{\bigstar}
}

\newcommand{\specialcell}[2][l]{%
  \begin{tabular}[#1]{@{}l@{}}#2\end{tabular}}

\newenvironment{shadequote}%
{\begin{snugshade}\begin{quote}}
{\hfill\end{quote}\end{snugshade}}

\definecolor{shadecolor}{rgb}{0.9,0.9,0.9}

\begin{document}
\title{Cross-Media Similarity Evaluation for \\ Web Image Retrieval in the Wild}

\author{Jianfeng~Dong,~Xirong~Li, and~Duanqing~Xu
\thanks{Manuscript received September 28, 2016; revised May 1, 2017 and November 25, 2017; accepted January 2, 2018. 
This work was supported by the National Science Foundation of China (No. 61672523) and the Key Scientific Research Base for Digital Conservation of Cave Temples (Zhejiang University),~State Administration for Cultural Heritage.
The associate editor coordinating the review of this manuscript and approving it for publication was Prof. Benoit HUET 
(\textit{Corresponding author: Xirong Li}).}
\thanks{J. Dong and D. Xu are with the College of Computer Science and Technology, Zhejiang University, Hangzhou 310027, China (e-mail: danieljf24@zju.edu.cn; xdq@zju.edu.cn).}
\thanks{X. Li is with the Key Lab of Data Engineering and Knowledge Engineering, School of Information, Renmin University of China, Beijing 100872, China (e-mail: xirong.li@gmail.com). }
}

\markboth{IEEE Transactions on Multimedia ,~Vol.~?, No.~?, ?~2018}%
{Dong \MakeLowercase{\textit{et al.}}: Cross-Media Similarity Evaluation for Web Image Retrieval in the Wild}

\maketitle

\begin{abstract}
In order to retrieve unlabeled images by textual queries, cross-media similarity computation is a key ingredient. Although novel methods are continuously introduced, little has been done to evaluate these methods together with large-scale query log analysis. Consequently, \emph{how far have these methods brought us in answering real-user queries} is unclear. Given baseline methods that use relatively simple text/image matching, \emph{how much progress have advanced models made} is also unclear. This paper takes a pragmatic approach to answering the two questions. Queries are automatically categorized according to the proposed query visualness measure, and later connected to the evaluation of multiple cross-media similarity models on three test sets. Such a connection reveals that the success of the state-of-the-art is mainly attributed to their good performance on visual-oriented queries, which account for only a small part of real-user queries. To quantify the current progress, we propose a simple \emph{text2image} method, representing a novel query by a set of images selected from large-scale query log. Consequently, computing cross-media similarity between the query and a given image boils down to comparing the visual similarity between the given image and the selected images. Image retrieval experiments on the challenging Clickture dataset show that the proposed text2image is a strong baseline, comparing favorably to recent deep learning alternatives.
\end{abstract}


\IEEEpeerreviewmaketitle

\section{Introduction} \label{sec:intro}

%
%
%
%

\IEEEPARstart{S}{ince} the early 1990s how to retrieve \emph{unlabeled} images by textual queries has been a grand challenge in multimedia retrieval, and remains hot to this day \cite{enster2008,yang2008harmonizing,Kang2015Learning,arXiv:1503.08248}.
In order to understand and exploit the interplay between visual content, textual query and user behaviors, web image retrieval demands multi-modal approaches \cite{Zhuang2008Mining,kofler2014intent,He2016Cross} and thus makes it right at the heart of the multimedia field.
As image and query are two distinct modalities, 
a cross-media similarity metric that effectively reflects image-query relevance is essential.

%
\begin{figure}[tb!]
\centering\includegraphics[width=0.9\columnwidth]{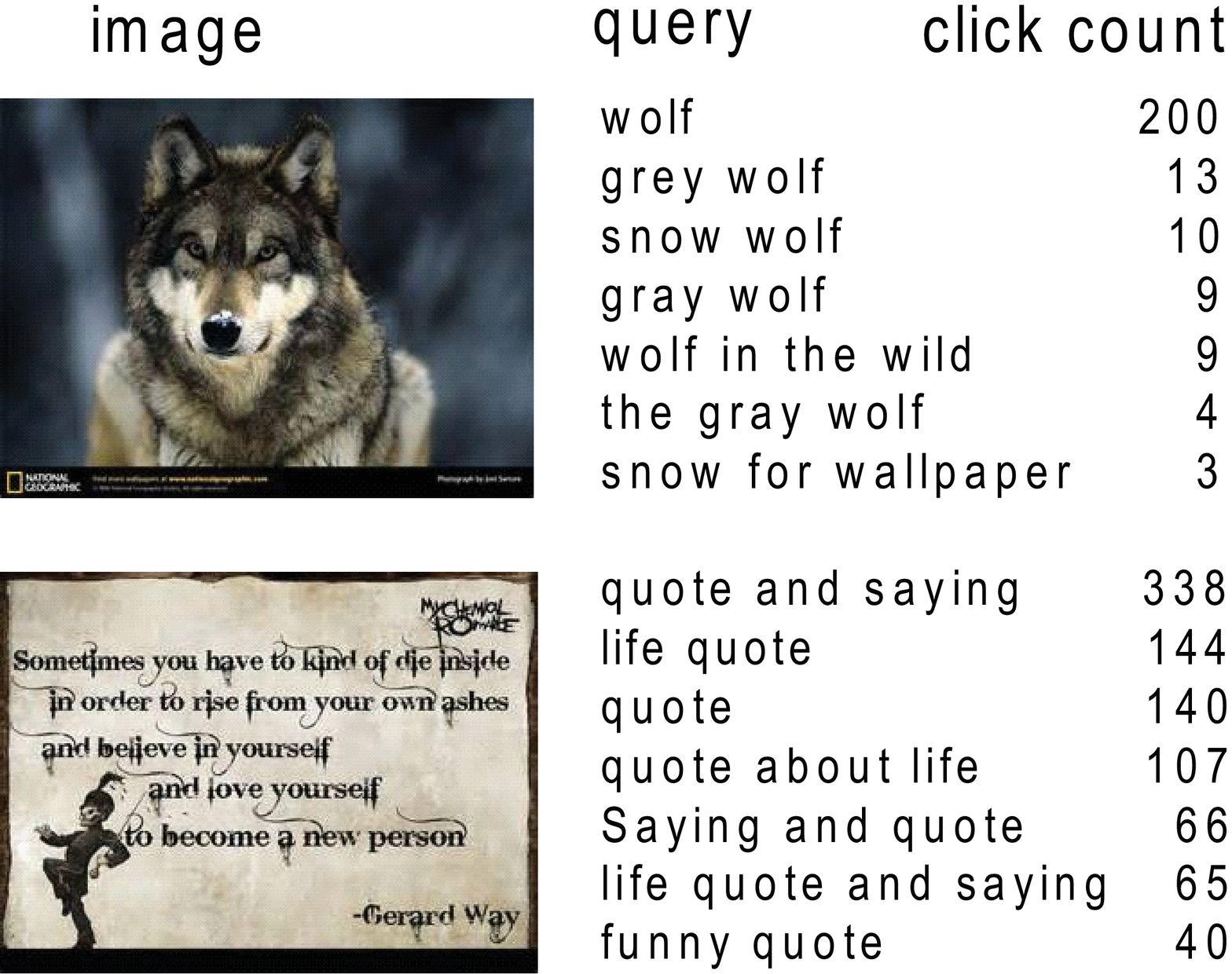}
\caption{\textbf{Image, query and their user-click count}, sampled from query log data of a web image search engine \cite{Hua2013Clickage}.
}
\label{fig:clickture_example}
\end{figure}

The key to cross-media similarity computation is to represent both images and queries in a common space \cite{nips13devise,Hua2014Cross,pami2014-pereira,yu2015learning, Deng2016Discriminative,gong2014multi}.
While the idea is simple, constructing a proper common space is non-trivial.
A desirable property is to let each of the many queries stay closer to images relevant w.r.t. the queries than irrelevant images.
In \cite{yu2015learning} for instance, Yu \etal employ deep neural network to embed images and queries into a joint latent space using large scale click-through logs.
Deng \etal \cite{Deng2016Discriminative} firstly perform dictionary learning to learn sparse and discriminative codes for images and queries respectively, and then map them into a common label space by linear transformation.

Depending on the choice of the common space, we see three lines of research.
One, represent images in a textual space by image auto-annotation \cite{Li2009Learning,Wang2010Towards,Ding2016Multi,Lu2016Tag} 
and consequently compute the cross-media similarity by text matching.
Two, represent queries in a visual space by learning visual templates of the queries \cite{Wu2007Learning,liu2011textual},
and compute the cross-media similarity by visual matching.
Third, also the main stream, build a common latent space, e.g., by maximizing correlation between relevant image-query pairs \cite{pami2014-pereira,wu2016learning, Deng2016Discriminative, Hua2014Cross}  or by minimizing a ranking based loss \cite{nips13devise,iccv2015-yaolearning, yu2015learning, He2016Cross}.
Given the progress described above, an important question then is 
\begin{shadequote}
 \emph{Q1. How far have cross-media similarity models brought us in answering user queries in the wild?} 
\end{shadequote}
By wild we mean any textual query one might submit to a generic web image search engine.
Such queries, due to their unconstrained nature, do not necessarily follow syntax, may contain typos, 
and more importantly, often lack correspondence to specific visual appearance, see Fig. \ref{fig:clickture_example}.
In spite of the rich literature, we argue that this question is largely untouched,
presumably due to the lack of related benchmark data.

For evaluating text-based image retrieval, 
we notice two types of benchmark data,
i.e., tag-based \cite{huiskes2010new,chua2009nus} and sentence-based \cite{hodosh2013framing,mscoco}.
MIRFlickr \cite{huiskes2010new} and NUS-WIDE \cite{chua2009nus} are two leading benchmark sets for tag-based image retrieval,
providing 14 and 81 test tags respectively.
These tags were chosen mainly because of their availability on Flickr.
In contrast to such single-tag queries, 
in Flickr8k \cite{hodosh2013framing} and MSCOCO \cite{mscoco}, 
a query is represented in the form of a sentence, which is meant for describing in brief main objects and scenes in an image.
The wiki text-image dataset \cite{rasiwasia2010new}  is an extreme case of sentence-based datasets, 
extending sentences to paragraphs. 
While searching for images by tags or sentences are of their own research interest,
they were chosen in favor of visual content analysis, 
and are unrepresentative of real-user queries.
Consequently,
to what extent conclusions established based on these benchmarks can be generalized to real-world image retrieval is unclear.

To promote research on web image retrieval by real-user queries, 
Hua \etal contribute Clickture-Lite, a click-through dataset down-sampled from one-year click log of the Bing image search engine \cite{Hua2013Clickage}.
This set contains one million images, 11.7 million distinct queries, 
and 23 million triads of (query, image, \emph{click}), where \emph{click} is the accumulated amount of user clicks a specific image has received with respect to a given query.
Although there are a growing amount of works on exploiting Clickture-Lite \cite{wu2013search,icme14uq,pan2014click2,liu2015click,yu2015learning,iccv2015-yaolearning,wu2016learning}, 
they focus on investigating the possibility of learning image retrieval models from the click-through data.
There lacks a systematic study that reveals what kind of queries the state-of-the-art can now handle.
Moreover, somewhat surprisingly, we observe that while more sophisticated models are introduced, 
they are hardly compared to simple approaches that compute cross-media similarity using relatively straightforward image-to-text \cite{pan2013image} or text-to-image mappings \cite{fang2013towards,wang2014france,dong2015fusion}.
Another surprising fact is that while statistical significance tests are the default for text retrieval evaluation \cite{cikm07-msmucker}, such tests are absent when assessing cross-media similarity models for web image retrieval. So the second question arises as 
\begin{shadequote}
\emph{Q2. How much progress have the advanced models significantly made when compared with the matching based baselines?}
\end{shadequote}

Previous works on image retrieval evaluation concentrate on content-based image retrieval (CBIR), where one searches for images by a given image~\cite{cvpr2005Shirahatti,mm2006jshen,Deselaers2008}. By contrast, we focus on retrieving \textit{unlabeled} images by free text. As such,  evaluating cross-media similarity between an image and a textual query is essential.

This paper takes a pragmatic approach to answering the above two questions, and contributes as follows:
\begin{itemize}
\item For answering \textit{Q1}, we introduce \emph{query visualness}, a quantifiable property for categorizing large-scale queries. Query visualness is found to be correlated with image retrieval performance. Such a connection helps us understand the merit and limit of the current models for addressing real-user queries. 
\item For answering  \textit{Q2}, we propose a simple  \textit{text2image} model, inspired by \cite{fang2013towards,wang2014france} but redesigned to better exploit click-through data for cross-media similarity computation. We present a systematic evaluation, comparing three advanced models with simple ones on three test sets, \ie MIRFlickr \cite{huiskes2010new} for single-tag queries and Clickture-dev \cite{Hua2013Clickage} and IRC-MM15-test\footnote{\url{http://research.microsoft.com/en-US/projects/irc/}} for real-user queries.
Consequently, we establish a new baseline for web image retrieval in the wild.
\end{itemize}

The techniques developed in this work have resulted in the winning entry in the Microsoft Image Retrieval Challenge at ACMMM’2015~\cite{dong2015fusion}.
Compared to our conference paper \cite{dong2015fusion} which aims in improving image retrieval performance by model fusion, this work goes one step further, trying to interpret the current progress on cross-media similarity computation for web image retrieval. This requires us to extend \cite{dong2015fusion} by including more models and more datasets to evaluate the effectiveness of the varied models for answering queries of varied difficulties. More importantly, to understand the gap between what the state-of-the-art can provide and what real users search for, such an evaluation has to be linked to automated query analysis. To this end, we introduce query visualness based analytics. All this results in a number of novel findings that are not covered by the conference paper. Source code is available at \url{https://github.com/danieljf24/cmrf}.

\section{Related Work} \label{sec:related}

This work is an endeavor towards quantifying progress on cross-media similarity computation for web image retrieval. So we first clarify our novelty in the context of image retrieval evaluation in general. As our research is at the crossroad of cross-media similarity computation and query log analysis, we then review recent progress in these two fields.

\subsection{Image Retrieval Evaluation}

Previous efforts on image retrieval evaluation focus on CBIR, where a user query is represented by a specific image.  In an early work \cite{cvpr2005Shirahatti}, Shirahatti and Barnard introduce a system for making grounded comparisons of different CBIR systems, using Corel images as their dataset. 
Shen and Shepherd \cite{mm2006jshen} propose a stratified sampling based approach for evaluating CBIR systems, providing both efficient and statistically-sound performance evaluation. 
Deselaers \etal \cite{Deselaers2008} present an experimental comparison of a large number of different low-level image descriptors.
As the similarity between the query image and the images being retrieved is computed directly in a visual feature space, cross-media similarity evaluation is out the scope of the above works.

To facilitate research on image retrieval by tags, Huiskes and Lew \cite{huiskes2008mir} provide MIRFlickr, a novel benchmark set collected from Flickr, followed by the NUS-WIDE dataset from Chua \etal \cite{chua2009nus}. 
Sun \etal \cite{jasist2011sun} conduct an empirical evaluation on tag-based social image retrieval, comparing the effectiveness of multiple tag-based ranking criteria. The study by Cheng \etal \cite{mmsy2016cheng} empirically investigates the effects of multiple information evidence on social image retrieval, where a query consists of a query
tag and an example image to facilitate different retrieval strategies. To attack the unreliability of social image tagging, Cui \etal \cite{mm2015cui} introduce a supervision step into the neighbor voting scheme \cite{tmm09-xirong} to make the neighbors reweighted towards optimizing the ranking performance of tag-based image retrieval, while Cui \etal \cite{jasist2015cui} improve neighbor voting by fusing multiple visual features. Besides tag-based image retrieval, we go a step further by considering real-user queries from a commercial web image search engine.

\subsection{Cross-Media Similarity Computation} \label{ssec:related-cm}

For embedding a textual query and an unlabeled image into a common space,
what matters are forms of the embeddings and objectives to be optimized.
So we review recent progress in these two aspects.

Regarding the forms, the main stream is to place an affine transformation either at the image side to project images into a bag-of-words space \cite{grangier2008discriminative} or at the query side to project queries into a visual feature space \cite{liu2011textual},
or per side a transformation to construct a latent space \cite{Bai2009Polynomial,pami2014-pereira,Deng2016Discriminative}. 
Depending on the choice of objectives, 
the embedding technique is known as Canonical Correlation Analysis (CCA) 
if one aims to maximize the correlation between embedding vectors of relevant pairs of query and image \cite{pami2014-pereira,pan2014click},
or as Polynomial Semantic Indexing (PSI)\cite{Bai2009Polynomial} if a marginal ranking loss is minimized.
In \cite{pan2014click2}, Pan \etal propose to minimize the distance of relevant pairs in the latent space,
with regularization terms to preserve the inherent structure in each original space.
A recent work by Yao \etal \cite{iccv2015-yaolearning} jointly uses CCA and PSI,
achieved by firstly finding a latent space by CCA and then re-adjusting the space to  incorporate ranking preferences from click-through data.

For the success of deep learning in computer vision and natural language processing,
we observe an increasing use of such techniques as an alternative to the affine transformation.
In \cite{yu2015learning}, for instance, Yu \etal use a deep Convolutional Neural Network (CNN) for image embedding,
while at the same time keep the transformation at the query side.
He \etal employ two CNNs for image and query embedding respectively \cite{He2016Cross}.
Frome \etal in their DeViSE model use a pre-trained word2vec for query embedding \cite{nips13devise}.
In a follow-up work, Norouzi \etal employ word2vec for both query and image embeddings \cite{Norouzi2013Zero}.
Bai \etal use a CNN that outputs a bag-of-words vector for an input image \cite{bai2014bag}, with no need of query embedding.
Wu \etal \cite{wu2016learning}  employ a graph-based representation learning algorithm to incorporate implicit connections among the click-through logs,
with the objective to minimize the negative log-likelihood.
While the models are becoming more sophisticated, 
the insight into the problem remains limited due to the lack of a joint analysis about the retrieval performance and properties of queries.
Moreover, there is hardly any comparison between these advanced models and naive ones based on simple text/image matching.

\subsection{Image Query Log Analysis} \label{ssec:related-log-analysis}

Existing works on image query log analysis focus on analyzing user search behavior \cite{pu-failed-query,hollink-three-level,ipm2001-excite},
characterized in several aspects such as what terms are used, how are they distributed, and how many terms per query.
In an early work \cite{ipm2001-excite}, 
Goodrum and Spink report that compared to text retrieval, users submit relatively few terms to specify their image information needs on the web.
This observation is confirmed by a more recent study by Hua \etal \cite{Hua2013Clickage}, reporting that around 83\% queries consist of two to five words.
To categorize words that ordinary users used for image description and for keyword-based image retrieval,
Hollink \etal propose a three-level abstraction \cite{hollink-three-level},
i.e., nonvisual, conceptual and perceptual.
These three levels correspond to information that cannot be derived from the visual content alone,
information about the semantics of the image,
and visual properties such as color, shape and texture.
Their study suggests that people tend to use more specific terms and less abstract and perceptual terms for image retrieval than for image description.
After analyzing some query log data of a local image search engine, Pu \cite{pu-failed-query} 
finds that failed queries are longer, more distinct and unique than successful queries.
These works provide good insights into image search behavior.
However, query categorization is conducted by hand, thus inapplicable for large-scale query log analysis.
Moreover, none of the above analysis is performed in conjunction with an evaluation of image retrieval models.

\section{Our Approach} \label{sec:approach}

\subsection{Cross-Media Similarity Evaluation} \label{ssec:methods}

Given an unlabeled image $x$ and a textual query $q$, 
we aim to construct a real-valued function $f(x,q)$ that computes their cross-media similarity.
To simplify the notation, $x$ also indicates a $d_i$-dimensional visual feature vector extracted from the image, while $q$ represents a $d_t$-dimensional bag-of-words vector extracted from the query. 
Apparently, the image feature $x$ and the query feature $q$ reside in different feature spaces, so they are not directly comparable.
We need to find a common space to represent them so that the similarity can be computed.
Similar to previous works \cite{pan2014click2,yu2015learning,dong2015fusion,liu2015click}, 
we build the common space by learning from large-scale click-through data, 
denoted as $\mathcal{D} = \{(image,query,click)\}$.

As discussed in Section \ref{ssec:related-cm}, the main stream is to implement the common space via varied semantic embedding techniques. 
More formally, suppose that the common space has a dimensionality of $d_c$,
with $d_c \le \min\{d_i,d_t\}$ typically.
We look for two transformations, $\phi_i(x):\mathbb{R}^{d_i}\rightarrow \mathbb{R}^{d_c}$ and $\phi_t(q):\mathbb{R}^{d_t}\rightarrow \mathbb{R}^{d_c}$, that embed $x$ and $q$ into the common space respectively.
Consequently, $f(x,q)$ can be expressed in terms of $\phi_i(x)$ and $\phi_t(q)$,
say, using an inner product.

Before delving into more sophisticated models for constructing $\phi_i(x)$ and $\phi_t(q)$,
we first describe in Section \ref{sssec:baselines} two baseline methods, 
i.e., image2text and text2image.
Later in Section \ref{sssec:latent} we depict three representative works on semantic embedding,
following the line of an increasing use of deep learning components.

\medskip

\subsubsection{Two Baselines} \label{sssec:baselines}

The following methods are considered as baselines, 
because they essentially compute cross-media similarity by straightforward image/text matching, without resorting to advanced learning techniques.

\medskip

\textbf{Baseline 1: image2text} \cite{pan2013image}.
For a given image $x$, we retrieve its $k$ nearest visual neighbors, 
denoted as $\{x_1,\ldots,x_k\}$, from $\mathcal{D}$.
The distance between two images is computed using the Euclidean Distance between their visual feature vectors.
Cross-media similarity between the image $x$ and a given query $q$ is computed as a weighted sum of the textual similarity between $q$ 
and queries associated with each of the  $k$ neighbor images. That is,
\begin{equation} \label{eq:i2t}
f_{i2t}(x,q) := \frac{1}{k}\sum_{i=1}^{k} sim(x,x_i) \cdot sim_{i2t}(x_i,q),
\end{equation}
where $sim(x,x')$ is an image-wise visual similarity, and
\begin{equation} \label{eq:simi2t}
sim_{i2t}(x_i,q) := \frac{1}{m_i} \sum_{j=1}^{m_i} sim(q, q_{i,j}) \cdot \log(click_{i,j}),
\end{equation}
where $m_i$ is the number of queries associated with the neighbor image $x_i$, $click_{i,j}$ is the click count $x_i$ received with respect to query $q_{i,j}$, and $sim(q,q')$ is a query-wise similarity.
Following \cite{pan2013image} we use the Jaccard similarity to realize $sim(q,q')$.
Viewing each query as a set of words, this similarity computes $sim(q,q')$ as $\frac{\left | q \cap q' \right |}{\left | q \cup  q' \right |}$, where $\left | q \cap q' \right |$ is the number of common words between $q$ and $q'$, and $\left | q \cup  q' \right |$ indicates the total number of unique words from the two queries. 
We have also tried the cosine similarity with queries represented by tf-idf vectors, and found the Jaccard similarity still better.
The contribution of query $q_{i,j}$ associated with the $i$-th neighbor image is weighted using its click count, see Eq. \ref{eq:simi2t}. A log scale is imposed on the count to improve stability.


\medskip

\textbf{Baseline 2: text2image}.
The text2image method reverses the mapping direction, projecting queries into the visual feature space.
Given a test query $q$, we first retrieve the top $k$ most similar queries,
denoted as $\{q_1,\ldots,q_k\}$, from $\mathcal{D}$.
The similarity between two queries is computed by the Jaccard similarity coefficient between their words, as done by image2text.
For a web image search engine, the same query can be submitted by distinct users as time goes.
Such a phenomenon is not uncommon as evidence shows newly-appeared queries only contribute to a relatively small proportion (around 11\%) of daily queries~\cite{liu2006automatic}. 
What an earlier user has clicked might also be relevant with respect to a later request. 
We leverage this intuition by setting $k$ to be 1 if the test query can be found in the provided query log $\mathcal{D}$.

For query representation, \cite{fang2013towards} uses all images associated with the neighbor queries.
We consider this design problematic because many of the images are irrelevant.
Alternatively, \cite{wang2014france} uses only the top five images most visually similar to the test image.
This strategy is also questionable because the test image itself can be irrelevant to the test query.
To represent the test query $q$ by a set of images more likely to be relevant, for each candidate image $x_i$ from the $j$-th neighbor query $q_j$, we estimate the relevance score between the test query and the candidate image by jointly considering the relevance between $x_i$ and $q_j$ and the relevance between $q_j$ and $q$, \ie
\begin{equation}
sim_{t2i}(x_i,q) := \log(click_{i,j}) \cdot sim(q,q_j).
\end{equation}
Accordingly, we sort all the candidate images in descending order by $sim_{t2i}(x_i,q)$, 
obtaining an ordered list of images $\{x_1,\ldots,x_{k'}\}$.
Note that for a candidate image associated with multiple queries, its $sim_{t2i}$ score is accumulated over the queries.
Consequently, the cross-media similarity is computed as a weighted sum of the visual similarity between $x$ 
and $\{x_1,\ldots,x_{k'}\}$. That is,
\begin{equation} \label{eq:t2i}
f_{t2i}(x,q) := \frac{1}{k'}\sum_{i=1}^{k'}  sim(x, x_{i}) \cdot sim_{t2i}(x_i,q).
\end{equation}

For both methods a standard text preprocessing is conducted: removing punctuation and lemmatizing words by NLTK~\cite{nltk}. Meaningless words in the context of image retrieval like `image' and `picture' and standard English stopwords are also removed. The parameter $k$ in image2text and text2image are empirically set to 50 and 30, respectively.

\medskip

\subsubsection{Three Semantic Embedding Models} \label{sssec:latent}

Among the many models, we opt to implement PSI \cite{Bai2009Polynomial} and
DeViSE \cite{nips13devise}, 
as they are key ingredients in varied methods for cross-media similarity computation.
In addition, we consider ConSE \cite{Norouzi2013Zero}, 
which is fully unsupervised and thus works across multiple datasets with ease.
The influence of these works is also demonstrated by the number of citations.

\medskip

\textbf{Model 1: PSI}.
It employs two affine transformations to project images and queries into a latent common space $\mathbb{R}^{d_c}$ as
\begin{equation}
\left\{ \begin{array}{rrl}
 \phi_i(x) & = & W_i x \\
  \phi_t(q) & = & W_t q 
  \end{array} \right.
\end{equation}
where $W_i \in \mathbb{R}^{d_c \times d_i} $ and $W_t \in \mathbb{R}^{d_c \times d_t}$ are trainable matrices.
The cross-media similarity is computed as a dot product between the embedding vectors,
\begin{equation} \label{eq:psi}
f_{psi}(x,q) := (W_i x)^T  (W_t q).
\end{equation}
The two matrices are optimized by minimizing a marginal ranking loss.
Concretely, we construct a large set of triplets $\mathcal{T}=\{(q, x^+, x^-) \}$ from $\mathcal{D}$,
where $x^+$  and $x^-$ indicate images relevant and irrelevant with respect to $q$.
The loss function is defined as 
\begin{equation} \label{eq:psi_loss}
L_{psi} := \sum_{(q, x^+, x^-) \in \mathcal{T}} max(0, 1 - f_{psi}(x^+,q) + f_{psi}(x^-,q)).
\end{equation}
We minimize $L_{psi}$ using stochastic gradient descent with a mini-batch size of 100.
In addition, we use an exponentially decaying learning rate, found to be useful for large-scale optimization \cite{hinton2012improving}.
Since PSI requires a predefined query vocabulary, 
we follow \cite{pan2014click,bai2014bag,yu2015learning}, 
preserving up to 50k most frequent words in the training data.

\medskip

\textbf{Model 2: DeViSE}.
The main difference between PSI and DeViSE is that the latter replaces the linear transformation $W_t$ by a pre-trained word2vec model to obtain $\phi_t(q)$.
Since the training process of word2vec is highly scalable and efficient, it builds embedding vectors for millions of words with ease.
Therefore, the size of the query vocabulary DeViSE can handle is much larger than that in PSI.

The embedding vector of a query $q$ is obtained by mean pooling over each word in the query:
\begin{equation} \label{eq:mean-pooling}
\phi_t^{devise}(q) :=\frac{1}{|q|}\sum_{w \in q} v(w),
\end{equation}
where $v(w)$ corresponds to the embedding vector of each word, and $|q|$ is the query length.
Recent studies report that word2vec trained on many Flickr tags better captures visual relationships than its counterpart learned from web documents \cite{li2015zero,mm2015emoji}. We follow such a tactic, training a 200-dimensional word2vec model on user tags of 25M Flickr images using the skip-gram algorithm \cite{word2vec}.

The DeViSE version cross-media similarity is computed as 
\begin{equation} \label{eq:devise-sim}
f_{devise}(x,q) := (W_i x)^T  \phi^{devise}_t(q).
\end{equation}
Due to the use of word2vec, DeViSE only needs to train the image transformation matrix $W_i$,
which is optimized in a similar way as PSI.
For a fair comparison, the dimension of PSI's common space is also set to 200.


\medskip

\textbf{Model 3: ConSE}.
Compared to DeViSE, ConSE goes one step further by employing a deep image classification model to embed the image to the word2vec space.
For a given image, a pre-trained CNN is used to predict the top $m=10$ most relevant labels, denoted as $\{y_1,\ldots,y_m\}$.
The image embedding vector is obtained as a convex combination of the embedding vectors of the labels, i.e.,
\begin{equation} \label{eq:conse}
\phi^{conse}_i(x) := \frac{1}{Z} \sum_{i=1}^m p(y_i|x) \cdot v(y_i),
\end{equation}
where $p(y_i|x)$ is the relevance score of $y_i$ given $x$, and $Z = \sum_{i=1}^m p(y_i|x)$ is a normalization factor.
The cross-media similarity is computed as cosine similarity in the word2vec space, i.e.,
\begin{equation} \label{eq:conse-sim}
f_{conse}(x,q) := cosine(\phi^{conse}_i(x),\phi^{devise}_t(q)).
\end{equation}
In contrast to the previous two models, ConSE is fully unsupervised. 
Its effectiveness relies on the quality of the top predicted labels for describing the image content.

Table \ref{tab:methods} summarizes the main properties of the two baselines and the three advanced models. The previous text preprocessing is conducted as well for the advanced models.

\begin{table} [tb!]
\renewcommand{\arraystretch}{1.2}
\caption{\textbf{Main properties of the five models} implemented in this work.}
\label{tab:methods}
\centering \scalebox{0.9}{
\begin{tabular}{@{}ll  l l l @{}}
\toprule
%
\textbf{Model}  && $\phi_i(x)$ & $\phi_t(q)$ & $f(x,q)$ \\
\cmidrule{1-1} \cmidrule{3-5} 
image2text \cite{pan2013image}  && bag-of-words &  $q$           & Eq. (\ref{eq:i2t}) \\
text2image (\emph{this work})   && $x$          & visual feature & Eq. (\ref{eq:t2i}) \\ 
PSI  \cite{Bai2009Polynomial}        && $W_i x$      & $W_t q$        & $\phi_i(x)^T \phi_t(q)$  \\
DeViSE \cite{nips13devise}      && $W_i x$      & word2vec       & $\phi_i(x)^T \phi_t(q)$ \\
ConSE  \cite{Norouzi2013Zero}      && word2vec     & word2vec       & $cosine(\phi_i(x),\phi_t(q))$     \\ 
\bottomrule
\end{tabular}
}
\end{table}

\medskip

\subsubsection{Cross-media Similarity Fusion} \label{sssec:fusion}
Since the above models compute $f(x,q)$ by distinct mechanisms, 
their output may complement each other.
To investigate whether combining them helps, we investigate cross-media similarity fusion.

For a given image-query pair, let $\{f_i(x,q)|i=1,\ldots,d\}$ be cross-media similarity scores computed by $d$ distinct models.
We consider the following late fusion strategy, for its simplicity and flexibility to employ a number of off-the-shelf learning to rank techniques:
\begin{equation} \label{eq:fusion}
f_\Lambda(x,q) := \sum_{i=1}^d \lambda_i \cdot  \sigma (f_i(x,q)),
\end{equation}
where $\Lambda=\{\lambda_i\}$ are weights to be optimized,
and $\sigma(\cdot)$ is a sigmoid function for rescaling the input.

Concerning the fusion weights, the simplest choice is to take uniform weights. 
This choice often works well in practice when the similarity functions to be fused are relatively close in terms of their performance and complementary to each other.
Once some ground truth data is provided, a range of learning to rank algorithms can be employed to find better weights.
We utilize Coordinate Ascent \cite{Metzler-LinearComination-IR07},
a greedy algorithm capable of directly optimizing (non-differentiable) performance metrics such as Average Precision and NDCG.

\subsection{Visualness based Query Log Analysis} \label{ssec:query-log}

It is clear that not all queries can be handled by a specific image retrieval model.
Knowing what kind of queries the model can address (or not) is beneficial, as it shows directions for improvement.
However, due to the complexity and diversity of real-user queries,
devising a comprehensive query categorizing scheme is extremely difficult, if not impossible.
As discussed in Section \ref{ssec:related-log-analysis},
existing categorization criteria such as query uniqueness \cite{pu-failed-query} and nonvisual/conceptual/perceptual \cite{hollink-three-level} 
are subjective and cannot be computed, making them inapplicable for image query log analysis at large-scale.

Since the state-of-the-art models compute  cross-media similarity on the basis of visual content, a desirable query property shall reflect the visual aspect of a given query. This connects to the degree of difficulty when a common user imagines what a relevant image would be for the given query. Visual concepts such as ``flower'' and ``bicycle'' are plausibly easier to be imagined than abstract concepts such as  ``design'' and ``saying''. Therefore, a query is more visual oriented if it contains more words corresponding to visual concepts.
%
Departing from this intuition, we propose to measure the visualness of a query by counting the proportion of its words that correspond to visual concepts.
Each query, by comparing its visualness score against a given threshold, can be automatically classified as either visual oriented or nonvisual.
This classification not only helps reveal the percentage of visual oriented queries in reality,
but also enables a fine-grained analysis of how a specific model responds to the two classes of queries.

Although there is no authoritative list of visual concepts,
we use the 21,841 concepts from ImageNet \cite{deng2009imagenet}, the largest labeled image collection in the public literature.
Each concept in ImageNet corresponds to a specific WordNet synset, describing visual objects and scenes in the world.
The concept is associated with one or more words or phrases.
Since individual words in a phrase, e.g., ``hot'' in ``hot dog'', are not necessarily visual,
a query or its fragment has to be fully matched with the phrase.
For a given query $q$, we define its visualness as
\small
\begin{equation} \label{eq:query-visualness}
\mbox{visualness}(q) := \frac{\mbox{\# of query words fully matched in ImageNet}}{\mbox{\# of query words}}.
\end{equation}
\normalsize

Table \ref{tab:query-visualness} presents some queries and their visualness scores.
The previous text preprocessing is applied before visualness analysis. 
Since ``design'' is a high-level notion lacking a consensus of imagery, the query ``tattoo design'' is considered less visual than the first three queries in Table \ref{tab:query-visualness}.
While being precise, a drawback of the vocabulary based measure is that it cannot handle visual concepts outside the ImageNet vocabulary. 
For instance, celebrity queries such as ``barack obama'' should have larger visualness scores, due to their correspondences to specific visual instances. However, they are not covered by ImageNet.
We argue that this limitation can be resolved to some extent by adding more domain-specific visual concepts.
An experiment regarding celebrity queries is provided in Section \ref{ssec:exp-query-analytics}.
Expanding the ImageNet vocabulary with a celebrity list allows ``barack obama'' to be covered, and consequently the visualness score of the query ``barack obama family'' will become 1. To handle fine-grained classes such as ``hot weather girl'', the vocabulary might be expanded automatically in a data-driven manner, by identifying new words suited for describing the visual content~\cite{tmm2010lu,mm2010sun}. A joint exploration of the vocabulary-based and data-driven approaches opens up possibilities for further improvement.

\begin{table} [tb!]
\renewcommand{\arraystretch}{1.2}
\caption{\textbf{Some queries and their visualness scores computed using Eq. (\ref{eq:query-visualness})}. 
Query words fully matched with specific ImageNet classes are marked out via square brackets. Strikethrough indicates stopwords removed by preprocessing.}
\label{tab:query-visualness}
\centering \scalebox{1}{
\begin{tabular}{@{}ll  l r@{}}
\toprule
%
Query && Visualness & Total click count \\
\cmidrule{1-1} \cmidrule{3-4}
{[flower]}                      && 1  & 220,747  \\
{[soccer ball]}               && 1  &  25,575  \\
{[dog]} \sout{and} {[cat]}         &&   1 & 3,423  \\
{[tattoo]} design             &&   0.500 & 59,854 \\   
barack obama {[family]}       &&   0.333 & 1,001 \\
hot weather {[girl]}          &&   0.333 &   31 \\
funny                           &&      0  & 578,659  \\
saying \sout{and} quote                &&      0  & 3,687 \\
\bottomrule
\end{tabular}
}
\end{table}

\begin{figure}[tb!]
\centering\includegraphics[width=0.9\columnwidth]{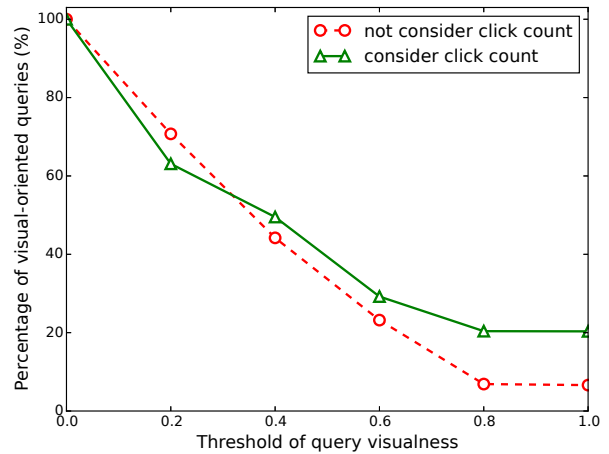}
\caption{\textbf{Percentage of visual-oriented queries in the Clickture-Lite dataset}. 
A query is classified as visual-oriented if its visualness score exceeds a given threshold.
Queries of larger visualness scores receive more user clicks,
indicating that they are better answered by the current image search engine.
}\label{fig:visual_realted}
\end{figure}

The dotted curve in Fig. \ref{fig:visual_realted} shows the percentage of visual-oriented queries in the Clickture-Lite dataset,
given varied thresholds. 
Queries with visualness scores exceeding 0.6 are less than 25\%.
This result suggests that even though the number of visual concepts we can learn (from the ImageNet) is bigger than ever, the learned concepts account for only a relatively small part of real-user queries. This situation will not improve due to the open-vocabulary aspect of queries submitted to a web image search engine.

Given queries of varied visualness, we further investigate how well the image search engine tackles them.
Because each query's click count reflects the chance of the search engine successfully returning relevant images for the query \cite{jasist2012-image-click}, we re-weight the query in terms of this value.
Given a specific threshold, the percentage of visual-oriented queries weighted with click count is computed as 
\begin{equation}
\frac{\sum_{q, visualness(q)> threshold} click_q}{\sum_{q'} click_{q'}},
\end{equation}
where $click_q$ is the accumulated click count of query $q$.
The updated percentage is shown as the solid curve in Fig. \ref{fig:visual_realted}, which goes above the original curve as the threshold increases, suggesting the current search engine better handles visual-oriented queries. This result also conforms to our daily experience with web image search.

\section{Evaluation} \label{sec:result}

To evaluate the effectiveness of the five models for answering queries of varied difficulty, we perform image retrieval by single-tag queries and by real-user queries. We choose MIRFlickr-test, a popular benchmark set for tag-based image retrieval \cite{huiskes2010new}. 
Popularized by the Microsoft Image Retrieval Challenge~\cite{Hua2013Clickage}, Clickture-dev is becoming a benchmark set for web image retrieval by real-user queries~\cite{pan2014click2,yu2015learning,bai2014bag,iccv2015-yaolearning,wu2016learning}. Nonetheless, the public availability of both images and ground-truth labels may unconsciously increase the chance of overfitting. Hence, we further include in our evaluation the test set from the Microsoft Image Retrieval Challenge at ACMMM'2015. We term the test set as IRC-MM15-test.
An overview of the three test sets is given in Table \ref{tab:testsets}. More details will be given in the following experiments.

\textbf{Evaluation criteria}. As Mean Average Precision (MAP) is commonly used on MIRFlickr \cite{arXiv:1503.08248}, we follow this convention. 
For Clickture-dev and IRC-MM15-test, we follow the MSR Image Retrieval Challenge protocol~\cite{Hua2013Clickage}, reporting Normalized Discounted Cumulated Gain (NDCG) at the rank of 25, \ie
\begin{equation} \label{eq:dcg25}
NDCG_{25} = 0.01757 \sum _{i=1}^{25} \frac{2^{rel_i - 1}}{log_2(i+1)} ,
\end{equation}
where $ rel_i = \{Excellent = 3, Good = 2, Bad = 0 \} $ is the ground truth label of the $i$-th ranked image with respect to the query,
while 0.01757 is a normalization factor to ensure that an ideal ranking will have an NDCG score of 1.
The overall performance is measured by averaging NDCG scores over the test queries.
As a sanity check, we report the performance of a random baseline on each test set, obtained by sorting images in terms of scores generated at random. On MIRFlickr-test, the random baseline has MAP of 0.0720, and NDCG\textsubscript{25} of 0.4702 and 0.4260 on Clickture-dev and IRC-MM15-test.

\textbf{Test of statistical significance}. 
We conduct a randomization test \cite{cikm07-msmucker}, with the null hypothesis that there is no performance difference in two image retrieval systems. Given a set of test queries and two retrieval systems, A and B, which have been evaluated per query. Let $diff$ be the absolute difference between the overall performance scores of the two systems. To check if the difference is caused by chance, for half of the test queries that are selected at random, their performance scores are switched between A and B. The absolute performance difference between A and B is re-computed accordingly and compared against $diff$. The trial is repeated $n$ times. The p-value of the randomization test is defined as the percentage of the trials having the new difference larger than $diff$. As suggested in \cite{cikm07-msmucker}, one can comfortably reject the null hypothesis if the p-value is smaller than 0.05. In other words, the performance difference between two image retrieval systems is considered statistically significant if the p-value produced by the randomization test is smaller than 0.05.

\begin{table} [tb!]
\renewcommand{\arraystretch}{1.2}
\caption{\textbf{Three test sets used in our experiments}. Models evaluated on MIRFlickr-test are learned from the MIRFlickr training set~\cite{huiskes2010new}, while models evaluated on Clicture-dev and IRC-MM15test are learned from the Clickture-Lite dataset~\cite{Hua2013Clickage}.}
\label{tab:testsets}
\centering \scalebox{0.87}{
\begin{tabular}{@{}l l l r r @{}}
\toprule
\textbf{Test set}        & \textbf{Source} & \textbf{Queries} & \textbf{Images} & \textbf{Image-query pairs} \\
\midrule
MIRFlickr-test  & Flickr   & 14 singel-tag queries    & 10,000   & 140,000 \\
Clickture-dev   & Web & 1,000 real-user queries & 79,665   & 79,926  \\
IRC-MM15test    & Web & 9,949 real-user queries & 147,346  & 579,122 \\
\bottomrule
\end{tabular}
}
\end{table}

\subsection{Experiment 1. Single-tag Queries} \label{ssec:exp-visual-query}

\textbf{Dataset}. The MIRFlickr set \cite{huiskes2010new} contains 25,000 Flickr images with ground truth annotations available for 14 tags such as `car', `dog' and `river'. Following the official partition, we use 10,000 images as the test set, termed as MIRFlickr-test. The remaining 15,000 images are used for model training. As all the test images are labeled with respect to the 14 test queries, this results in 14$\times$10,000=140,000 image-query pairs. 

\textbf{Visual features}. We employ three pre-trained CNNs, i.e., 
CaffeNet \cite{Jia2014Caffe}, VggNet \cite{simonyan2014very}, and GoogleNet \cite{szegedy2014going},
as their distinct network architectures may yield complementary visual features.
They were learned from examples of 1,000 ImageNet classes 
defined in the Large Scale Visual Recognition Challenge \cite{ILSVRCarxiv14}.
For CaffeNet and VggNet, we use the last fully connected layer, resulting in visual feature vectors of 4,096 dimensions.
For GoogleNet, we use its pool5 layer,
but replace the default $7 \times 7$ filter by a $4 \times 4$ filter to better capture spatial layouts.
This also results in a feature vector of 4,096 dimensions.
Note the three CNN models are separately used as the image classification component in ConSE.
For simplicity, the visual features are named after the corresponding CNNs.
CaffeNet is the default feature unless otherwise stated.

\textbf{Advanced models versus baselines}. 
The performance of different methods on MIRFlickr-test is summarized in Table \ref{tab:mir_map}.
All the methods are noticeably better than the random result.
PSI is at the leading position, followed by DeViSE.
Recall that their main difference lies in the underlying approach of query embedding.
The result suggests that for embedding single-tag queries, a task-specific transformation matrix is more suited than a word2vec model learned in advance.

\begin{table} [tb!]
\renewcommand{\arraystretch}{1.1}
\caption{\textbf{Performance of different models on answering single-tag queries on MIRFlickr-test}. 
Average weights are used for both feature-fusion and method-fusion.}
\label{tab:mir_map}
\centering \scalebox{0.9}{
\begin{tabular}{@{}ll  rrrrrrr@{}}
\toprule
%
Method && CaffeNet &&  VggNet &&  GoogleNet &&  \emph{Feature-fusion}\\
\cmidrule{1-1} \cmidrule{3-7}  \cmidrule{9-9}
random   &&  0.0720  &&  0.0720 &&  0.0720 &&  - \\ 
Upper bound       &&  1.0     &&  1.0    &&  1.0 &&  - \\ [4pt]
image2text        &&  0.4416  &&  0.4895 &&  0.4527 &&  0.5363 \\
text2image        &&  0.4414  &&  0.4930 &&  0.4753 &&  0.5226 \\
PSI               &&  0.4689  &&  0.5358 &&  0.5323 &&  \textbf{0.6544} \\
DeViSE            &&  0.4626  &&  0.5342 &&  0.5036 &&  0.6312 \\ 
ConSE             &&  0.3419  &&  0.3750 &&  0.3745 &&  0.4370 \\ [4pt]
\emph{Method-fusion}     &&  0.5643  &&  \textbf{0.6383} &&  0.5972 &&  0.6655 \\ 

\bottomrule
\end{tabular}
}
\end{table}

The above conclusion is further supported by the relatively lower performance of ConSE, which fully uses pre-trained models for both query and image embedding. While sharing the same word2vec model with DeViSE, at the image side ConSE counts on the 1K ImageNet classes to describe the visual content, which are too specific to represent the 14 test queries. Moreover, the advantage of one model over another is feature independent. For instance, PSI consistently outperforms DeViSE given all the three features.

Concerning the two baseline models, image2text with feature-fusion scores higher MAP of 0.5363 than text2image with MAP of 0.5226. Nonetheless, with p-value of 0.775 by randomization test, the difference is not statistically significant.

\begin{figure}[tb!]
\centering\includegraphics[width=0.85\columnwidth]{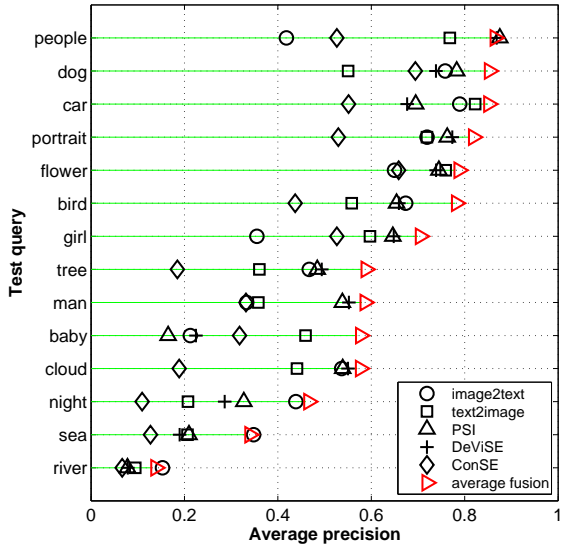}
\caption{\textbf{Per-query comparison on MIRFlickr-test}. Visual feature: VggNet.
Queries are sorted in terms of the performance of average-fusion.}\label{fig:mir_map_tag}
\end{figure}

\textbf{The influence of fusion}. 
As shown in Table \ref{tab:mir_map}, average-fusion, either along the line of methods (\emph{method-fusion}) or along the line of features (\emph{feature-fusion}), leads to significant improvements.
For instance, among all the fifteen combinations of methods and features, 
PSI + VggNet is the best. 
This run can be further improved by adding PSI + CaffeNet and PSI + GoogleNet, lifting MAP from 0.5358 to 0.6544.
A per-query comparison is given in Fig. \ref{fig:mir_map_tag},
where fusion performs the best for the majority of the queries.
Average fusion of all the fifteen combinations generates MAP of 0.6655.
Moreover, we employ Coordinate Ascent to optimize the weights on a set of 5k images sampled at random from the MIRFlickr training set.
With the learned weights, the performance can be further improved,  reaching MAP of 0.6772.

\subsection{Experiment 2. Real-user Queries} \label{ssec:exp-real-query}



\begin{figure}[tb!]
\centering\includegraphics[width=0.9\columnwidth]{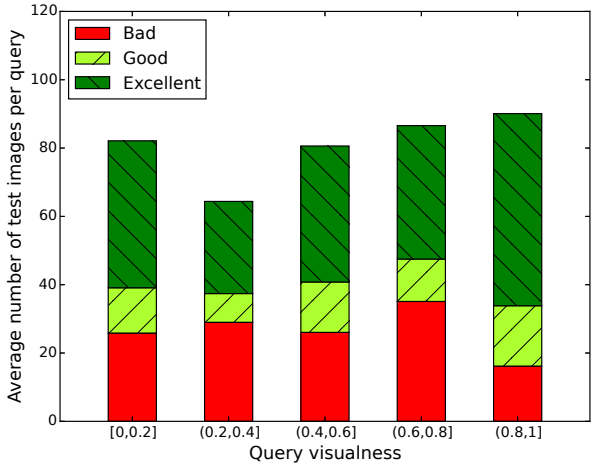}
\caption{\textbf{Average number of test images per query in Clickture-dev}.
Queries have been grouped in terms of their visualness scores.
  }\label{fig:pos_neg_3}
\end{figure}

\textbf{Test set 1: Clickture-dev}. 
This test set contains 1,000 real-user queries and 79,665 images. Though in theory there shall be 1000$\times$79,665 query-image pairs, only ground truth of 79,926 pairs are publicly available. Thus, image retrieval on Clickture-dev is to score and rank a subset of the images specific for each query.  Each pair is manually rated as \textit{Excellent}, \textit{Good} or \textit{Bad}, based on the relevance between image and query.
The distribution of each label is visualized in Fig. \ref{fig:pos_neg_3}. 

\textbf{Test set 2: IRC-MM15-test}. 
This test set contains 9,949 real queries and 579,122 query-image pairs. The set is much larger and more challenging than Clickture-dev.
As its ground truth is non-public, we submit our results to the task organizers and get performance scores back.

When tested on Clickture-dev and IRC-MM15-test, all the models use Clickture-Lite as the training data.


\textbf{Advanced models versus baselines}.
Table \ref{tab:irc_ndcg} shows the performance of the individual methods with varied features on Clickture-dev. 
Note that the performance upper bound is less than 1, 
because 609 of the 1,000 test queries have less than 25 images labeled as Excellent.
Again, all the methods beat the random result.
It is worth pointing out that retrieving images from an unconstrained collection of thousands of real-user queries is a grand challenge. Hence, although the performance divergence may appear to be relatively small (see also the performance reported in \cite{yu2015learning,wu2016learning}), the significance of the individual models shall not be underestimated.

\begin{figure}[tb!]
\centering\includegraphics[width=0.9\columnwidth]{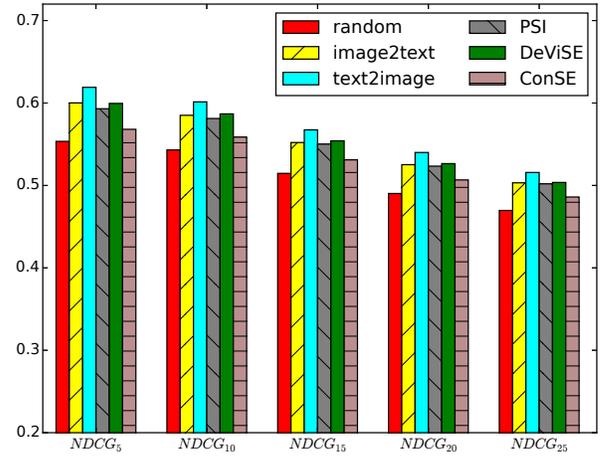}
\caption{\textbf{NDCG results at multiple ranks}. Test set: Clickture-dev. The text2image model performs the best.}\label{fig:ndcg_rank}
\end{figure}

When ranking the methods in terms of their performance, 
the resultant order differs much from that in Experiment 1.
The second baseline method, text2image, now takes the leading position,
followed by DeViSE, PSI, image2text and ConSE.
Fig. \ref{fig:ndcg_rank} shows NDCG results at multiple ranks, where text2image still performs the best.

In spite of its simplicity, text2image benefits from the large-scale query log data.
We find that for 42\% of the test queries, there exist precise matches in Clickture-Lite.
In these cases, cross-media similarity computation boils down to 
comparing a test image with several highly clicked images of the query.
As a dual form of text2image, image2text also compares images,
but works in a reverse order by finding similar images first.
This may incorrectly introduce irrelevant images and consequently propagate irrelevant words to the test images.
While such noisy words are not critical for single-tag queries (as in Experiment 1),
they affect complex queries. 
Consequently, we see from Table \ref{tab:irc_ndcg} that text2image consistently outperforms image2text for real-user queries.

\begin{table} [tb!]
\renewcommand{\arraystretch}{1.2}
\caption{\textbf{Performance of different models for answering real-user queries on Clickture-dev}. Average weights are used for both feature-fusion and method-fusion.}
\label{tab:irc_ndcg}
\centering \scalebox{0.9}{
\begin{tabular}{@{}ll  rrrrrrr@{}}
\toprule
%
Method && CaffeNet &&  VggNet &&  GoogleNet &&  \emph{Feature-fusion}\\
\cmidrule{1-1} \cmidrule{3-7}  \cmidrule{9-9}
random   &&  0.4702  &&  0.4702 &&  0.4702 &&  - \\
Upper bound       &&  0.6842  &&  0.6842 &&  0.6842 &&  - \\ [4pt]
image2text        &&  0.5032  &&  0.5025 &&  0.5043 &&  0.5055 \\
text2image        &&  0.5153  &&  0.5105 &&  0.5127 &&  \textbf{0.5149} \\ 
PSI               &&  0.5016  &&  0.5028 &&  0.5037 &&  0.5086 \\
DeViSE            &&  0.5033  &&  0.5020 &&  0.5070 &&  0.5099 \\
ConSE             &&  0.4861  &&  0.4878 &&  0.4837 &&  0.4882 \\ [4pt]
\emph{Method-fusion}    &&  0.5137  &&  \textbf{0.5145} &&  0.5142 &&  0.5177 \\ 


\bottomrule
\end{tabular}
}
\end{table}

\begin{figure}[tb!]
\centering\includegraphics[width=\columnwidth]{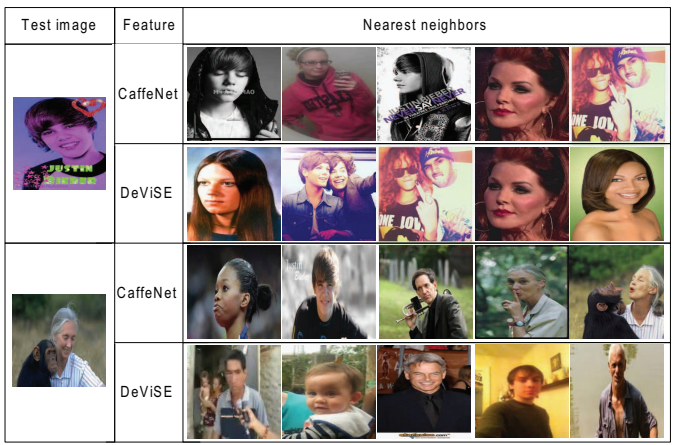}
\caption{\textbf{Top five similar images retrieved using the CaffeNet feature and the subspace feature by DeViSE, separately}. 
The former gets more images of the same person (Justin Bierber and Jane Goodall) as in the test images.}\label{fig:similiar-images}
\end{figure}

As for the semantic embedding models, 
in essence they aim to describe both images and queries by some latent topics.
While the topics provide higher level abstraction than visual features and bag-of-words features, 
and this is a wanted property for visual concept search,
the discrimination ability of both images and queries is inevitably reduced.
See Fig. \ref{fig:similiar-images} for instance.
On the other hand, by looking into the individual queries of Clickture-dev, we observe that many of the real-user queries are related to finding instances instead of categories, as exemplified in Table \ref{tab:search-results}.
Consider the query `ling simpson' for instance. While the true answer is about a female cartoon character, DeViSE retrieves images of real females.  Again, when coming to categorical queries such as `family photo' and `woman bicycle', DeViSE successfully find relevant images.
Therefore, despite their superior performance for single-tag queries which are visual concepts,
DeViSE and PSI are less effective for real-user queries.

\begin{table*} [tb!]
\renewcommand{\arraystretch}{1.2}
\caption{\textbf{Retrieval results by text2image and DeViSE}. 
Excellent, Good or Bad images are indicated by $\bigstar\bigstar$, $\bigstar\openbigstar$, and $\openbigstar\openbigstar$, respectively. }
\label{tab:search-results}
\centering
\scalebox{.85}{
\begin{tabular}{@{}| c | c | c | l | lllll |@{}}
\toprule

\textbf{Test query} & \textbf{Method} & \textbf{NDCG@25} & \textbf{Queries retrieval from Clickture-Lite} & \multicolumn{5}{c|}{\textbf{Images retrieval from Clickture-dev}}  \\
\midrule

\multirow{10}{*}{\specialcell{ 2001 ford \\ expedition part \\ \\visualness: 0}} & \multirow{5}{*}{text2image} &  \multirow{5}{*}{0.3597}  & 
\multirow{5}{*}{\specialcell{ 2001 ford expedition motor part \\  ford expedition part \\ 2001 ford expedition  \\ ford expedition part diagram \\  ford expedition part diagram}}
& \multirow{3}{*}{\raisebox{-0.5\totalheight}{\includegraphics[width=1.6cm,height=1.6cm]{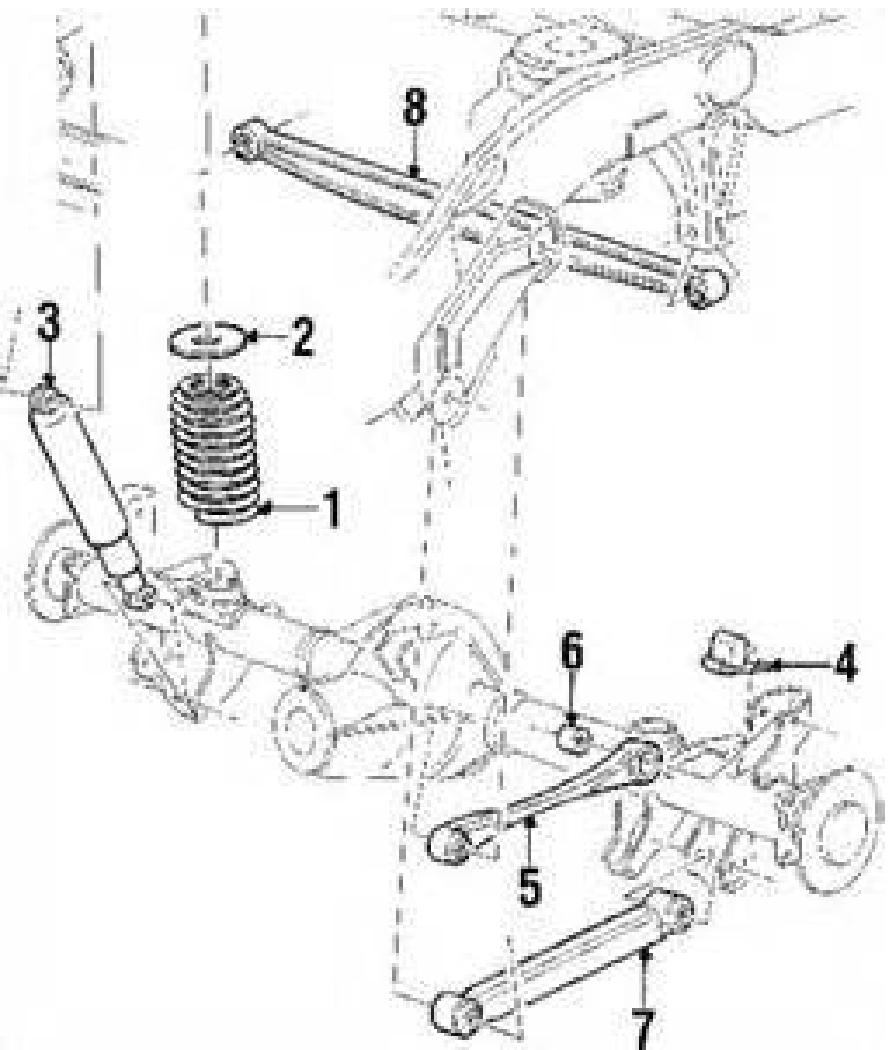}}} 
& \multirow{3}{*}{\raisebox{-0.5\totalheight}{\includegraphics[width=1.6cm,height=1.6cm]{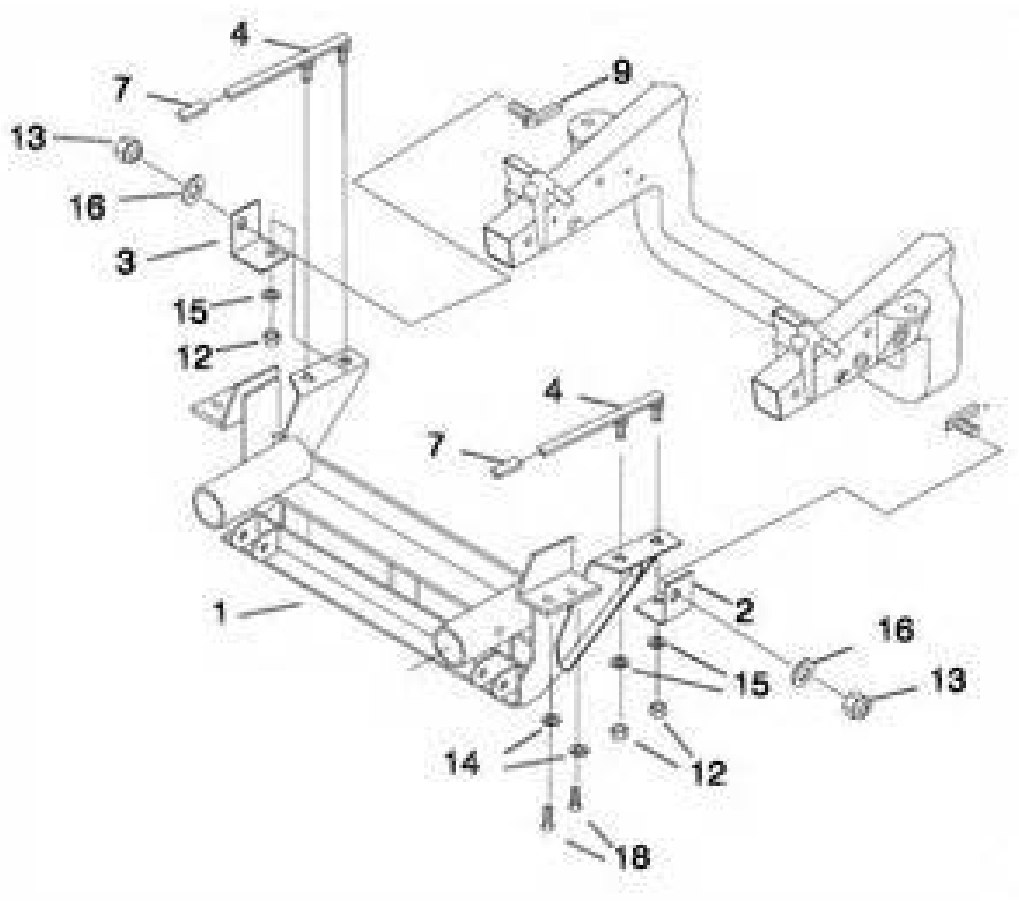}}} 
& \multirow{3}{*}{\raisebox{-0.5\totalheight}{\includegraphics[width=1.6cm,height=1.6cm]{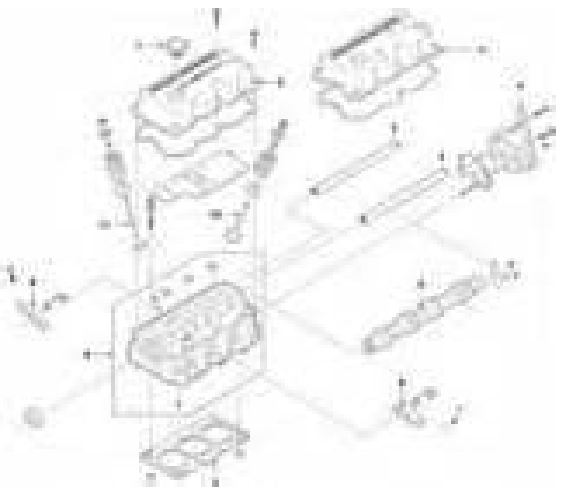}}} 
& \multirow{3}{*}{\raisebox{-0.5\totalheight}{\includegraphics[width=1.6cm,height=1.6cm]{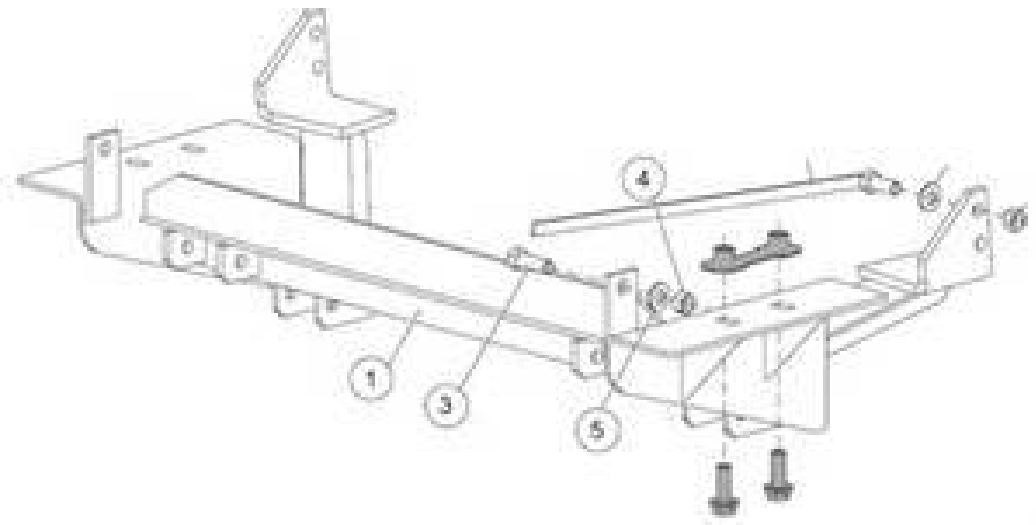}}} 
& \multirow{3}{*}{\raisebox{-0.5\totalheight}{\includegraphics[width=1.6cm,height=1.6cm]{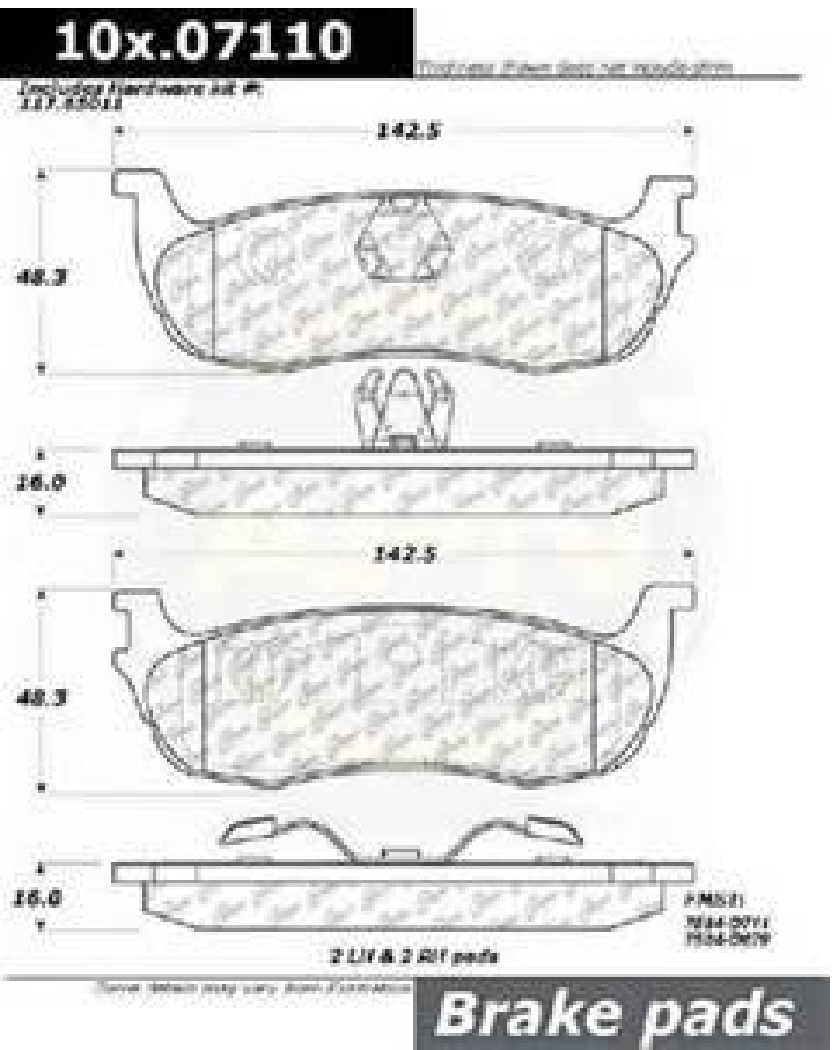}}}  \\
& & &  &&&&&\\
& & &  &&&&&\\ 
& & &  &&&&&\\ 
& & &  & $\bigstar\openbigstar$& $\bigstar\openbigstar$&$\openbigstar\openbigstar$& $\bigstar\openbigstar$&$\bigstar\openbigstar$\\
\cmidrule(l){2-9}
 & \multirow{5}{*}{DeViSE} &   \multirow{5}{*}{0.0}  & 
 \multirow{5}{*}{\specialcell{ 2002 ford expedition part \\ 2003 ford expedition part \\ 1999 ford expedition part \\ ford expedition 1999 part \\ 1997 ford expedition part }}
 & \multirow{3}{*}{\raisebox{-0.5\totalheight}{\includegraphics[width=1.6cm,height=1.6cm]{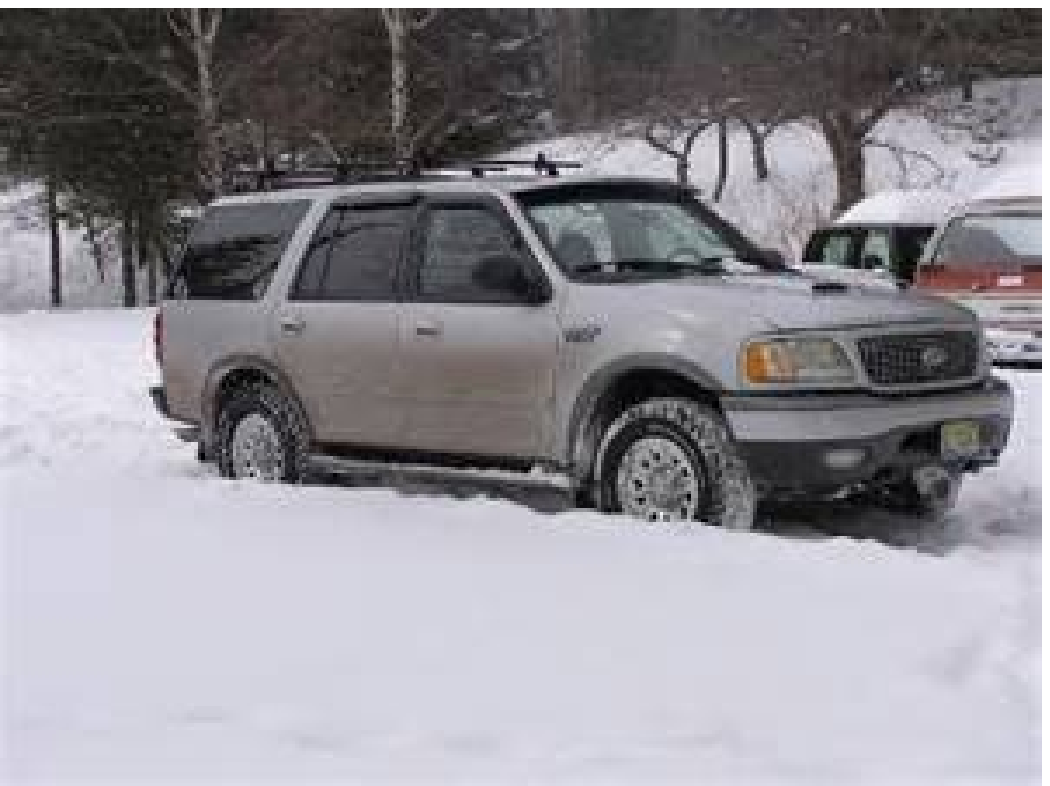}}} 
 & \multirow{3}{*}{\raisebox{-0.5\totalheight}{\includegraphics[width=1.6cm,height=1.6cm]{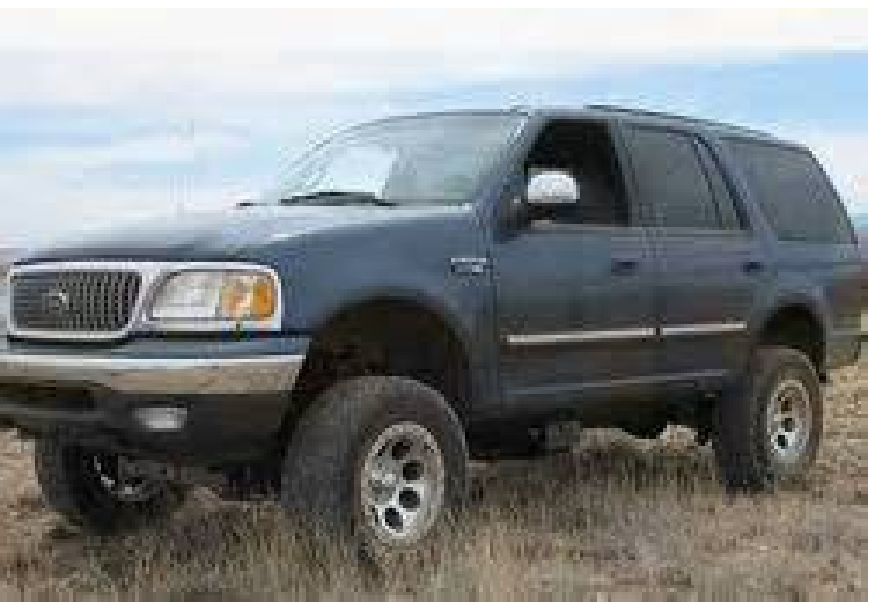}}} 
 & \multirow{3}{*}{\raisebox{-0.5\totalheight}{\includegraphics[width=1.6cm,height=1.6cm]{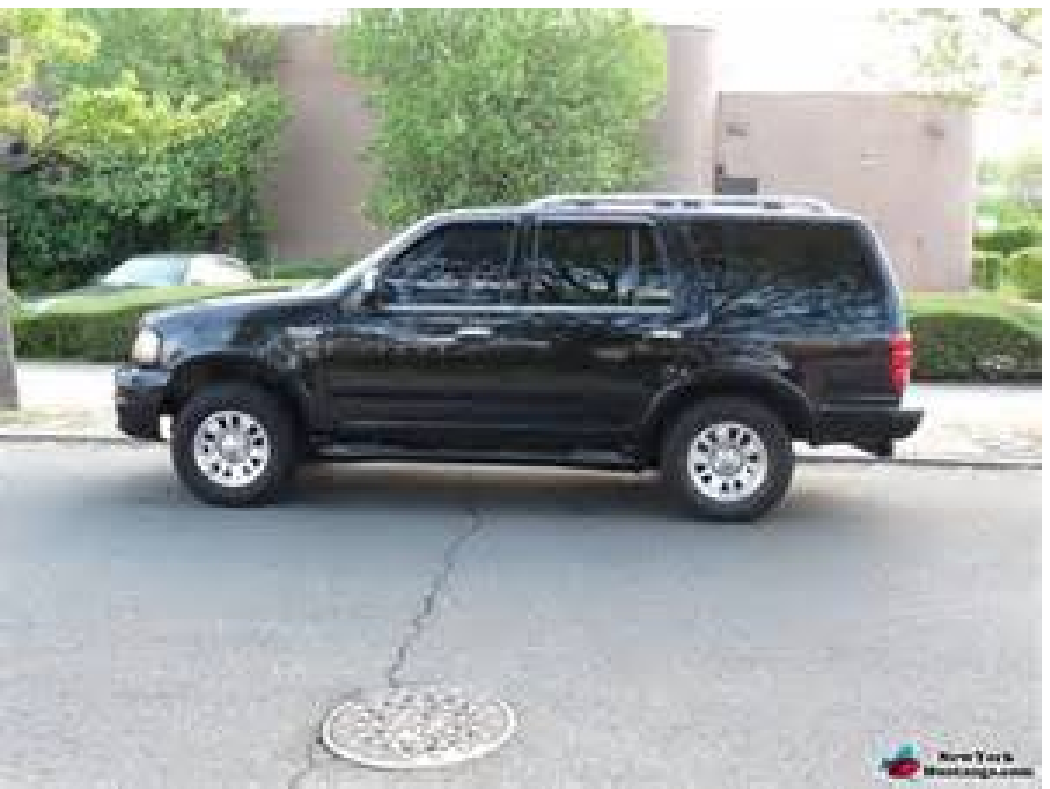}}}  
 & \multirow{3}{*}{\raisebox{-0.5\totalheight}{\includegraphics[width=1.6cm,height=1.6cm]{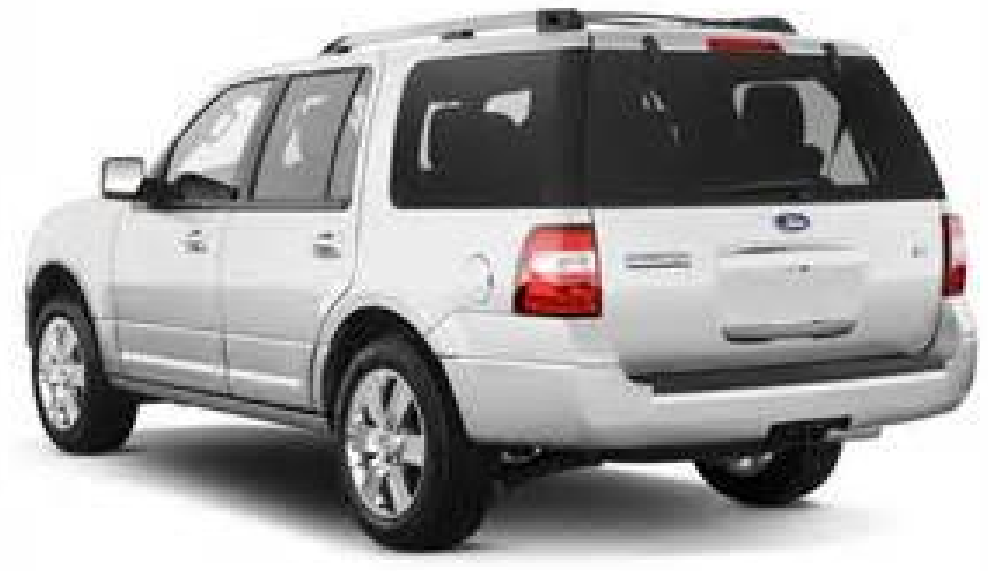}}} 
 & \multirow{3}{*}{\raisebox{-0.5\totalheight}{\includegraphics[width=1.6cm,height=1.6cm]{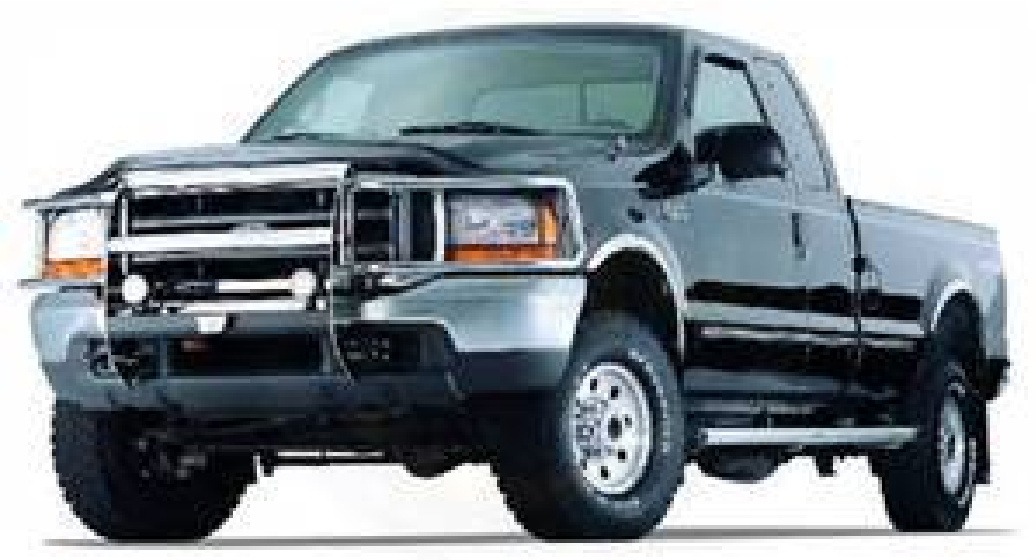}}}  \\
 & & &  &&&&&\\
 & & &  &&&&&\\ 
 & & &  &&&&&\\ 
 & & &  &$\openbigstar\openbigstar$&$\openbigstar\openbigstar$&$\openbigstar\openbigstar$&$\openbigstar\openbigstar$&$\openbigstar\openbigstar$\\
\midrule

\multirow{10}{*}{\specialcell{6v {[battery]} \\ small\\ \\visualness: 0.333}} & \multirow{5}{*}{text2image} &  \multirow{5}{*}{0.6206}  & 
\multirow{5}{*}{\specialcell{ 6v battery \\ small battery heater \\ 6v lantern battery \\ inside 6v battery  \\  small battery candle }}
& \multirow{3}{*}{\raisebox{-0.5\totalheight}{\includegraphics[width=1.6cm,height=1.6cm]{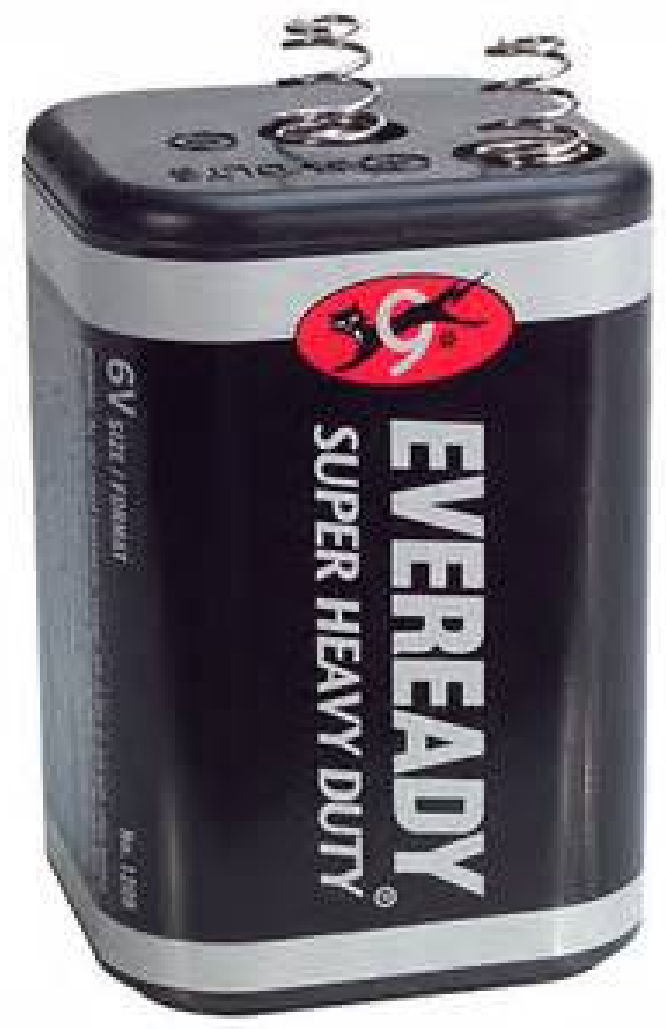}}} 
& \multirow{3}{*}{\raisebox{-0.5\totalheight}{\includegraphics[width=1.6cm,height=1.6cm]{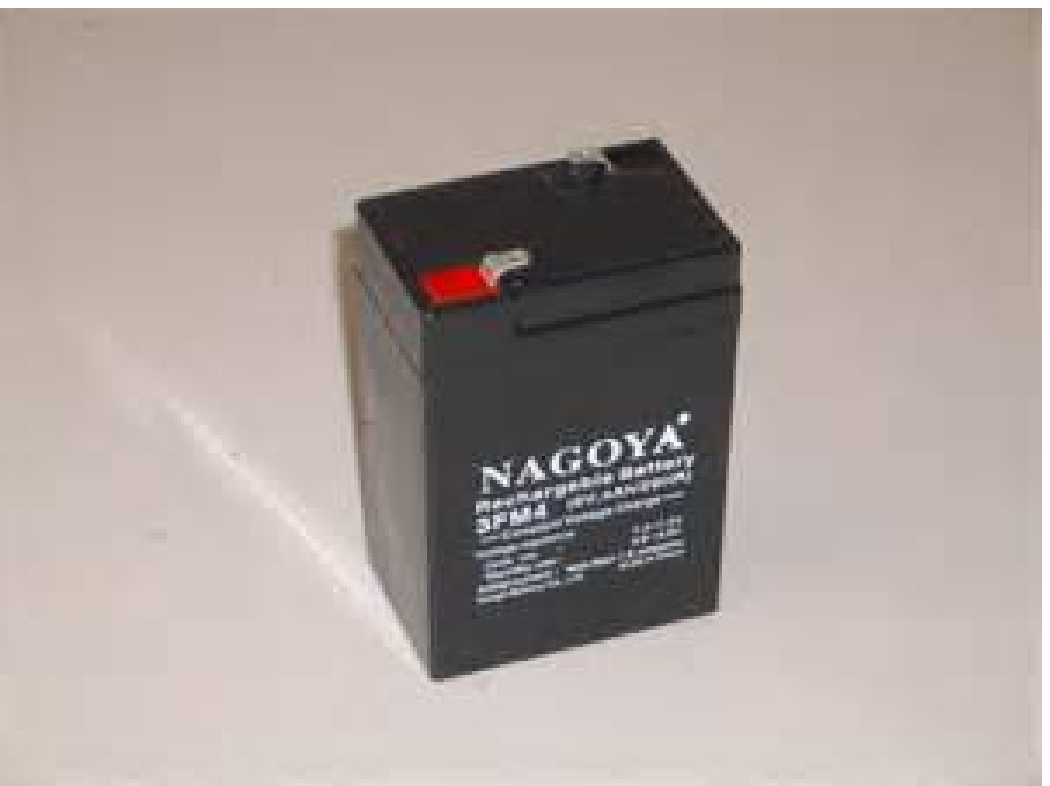}}} 
& \multirow{3}{*}{\raisebox{-0.5\totalheight}{\includegraphics[width=1.6cm,height=1.6cm]{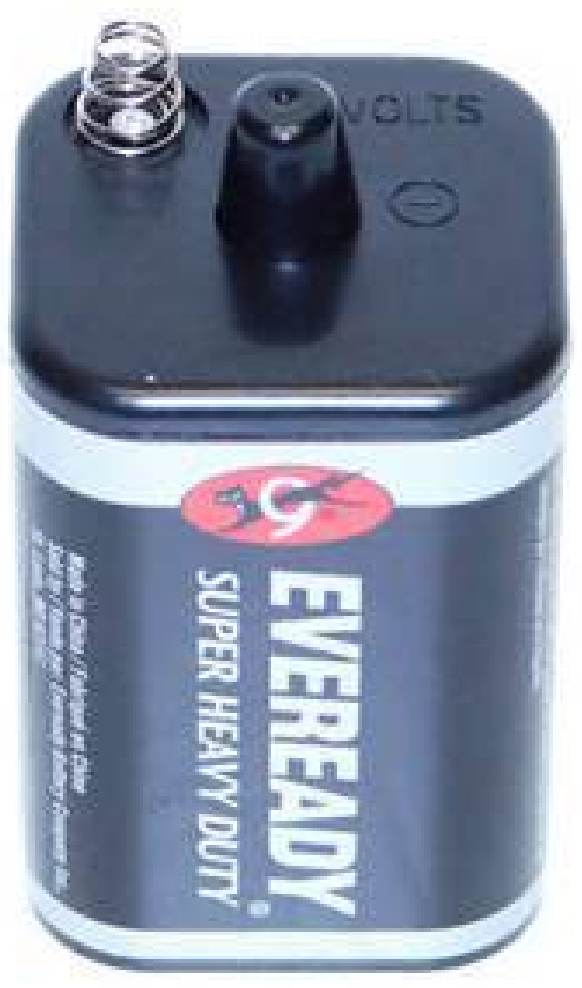}}} 
& \multirow{3}{*}{\raisebox{-0.5\totalheight}{\includegraphics[width=1.6cm,height=1.6cm]{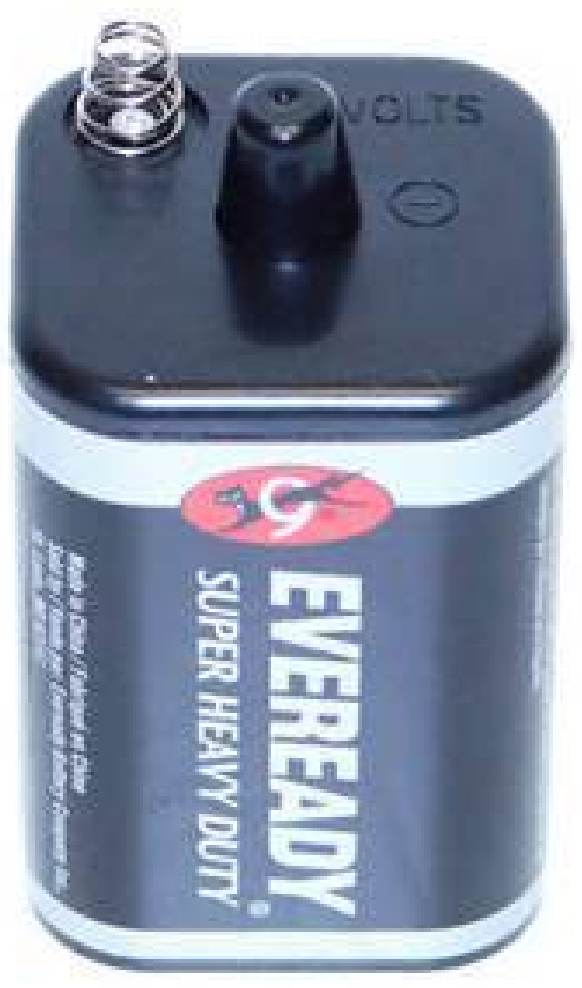}}} 
& \multirow{3}{*}{\raisebox{-0.5\totalheight}{\includegraphics[width=1.6cm,height=1.6cm]{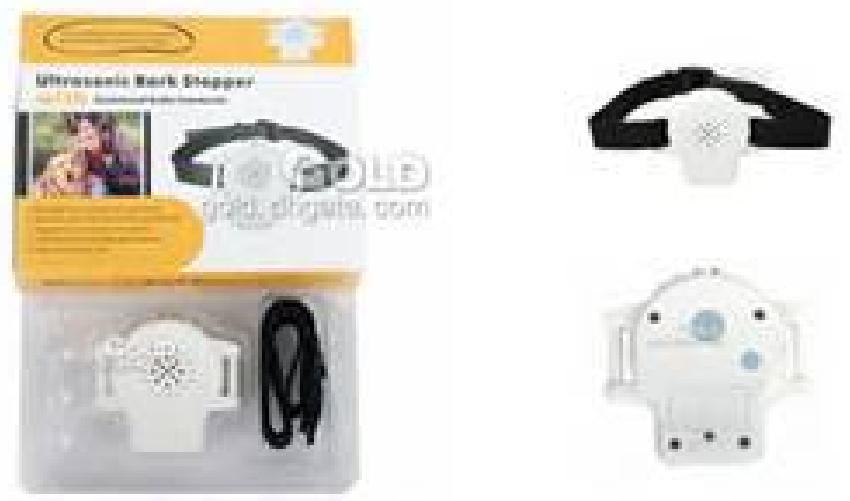}}}  \\ 
& & &   &&&&&\\   
& & &   &&&&&\\   
& & &   &&&&&\\ 
& & &  &$\bigstar\bigstar$& $\bigstar\bigstar$&$\bigstar\bigstar$& $\bigstar\bigstar$&$\openbigstar\openbigstar$\\
\cmidrule(l){2-9}
 & \multirow{5}{*}{DeViSE} &  \multirow{5}{*}{0.2155} & 
 \multirow{5}{*}{\specialcell{ 6v battery \\  inside 6v battery \\ small battery chargeing device \\ 6volt battery \\ labled battery }}
 & \multirow{3}{*}{\raisebox{-0.5\totalheight}{\includegraphics[width=1.6cm,height=1.6cm]{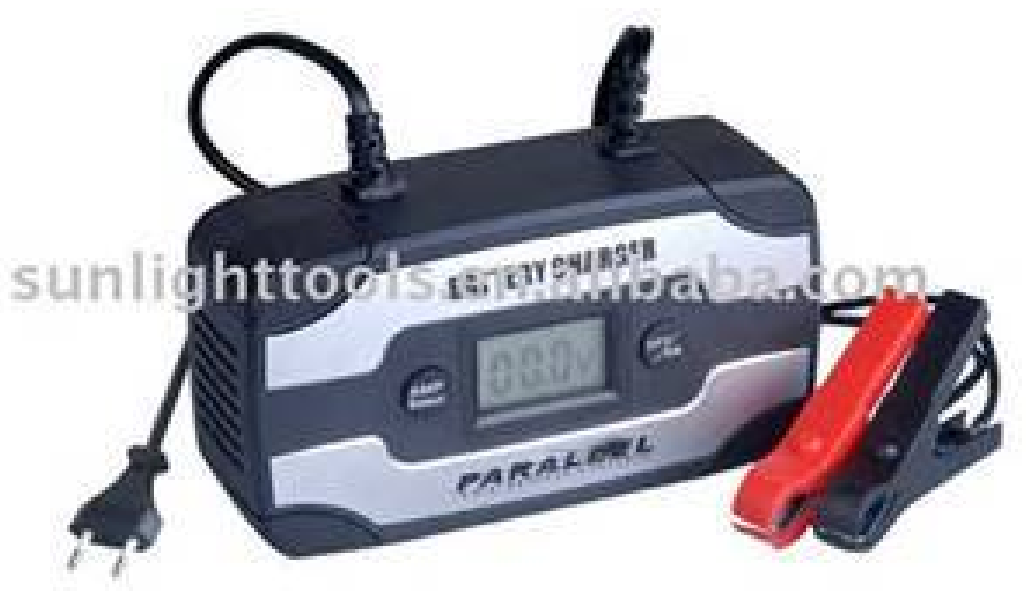}}} 
 & \multirow{3}{*}{\raisebox{-0.5\totalheight}{\includegraphics[width=1.6cm,height=1.6cm]{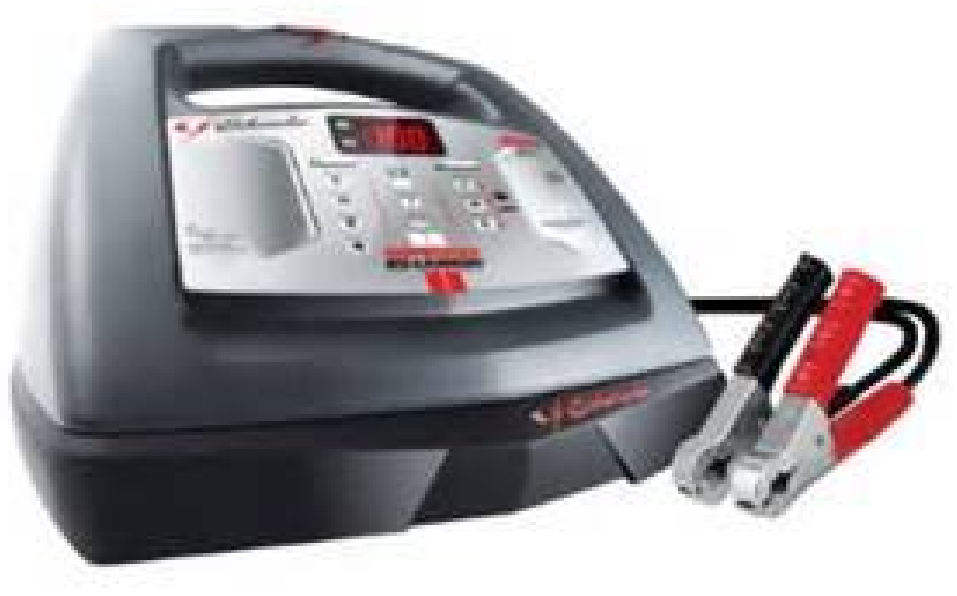}}} 
 & \multirow{3}{*}{\raisebox{-0.5\totalheight}{\includegraphics[width=1.6cm,height=1.6cm]{search_example/5224}}}  
 & \multirow{3}{*}{\raisebox{-0.5\totalheight}{\includegraphics[width=1.6cm,height=1.6cm]{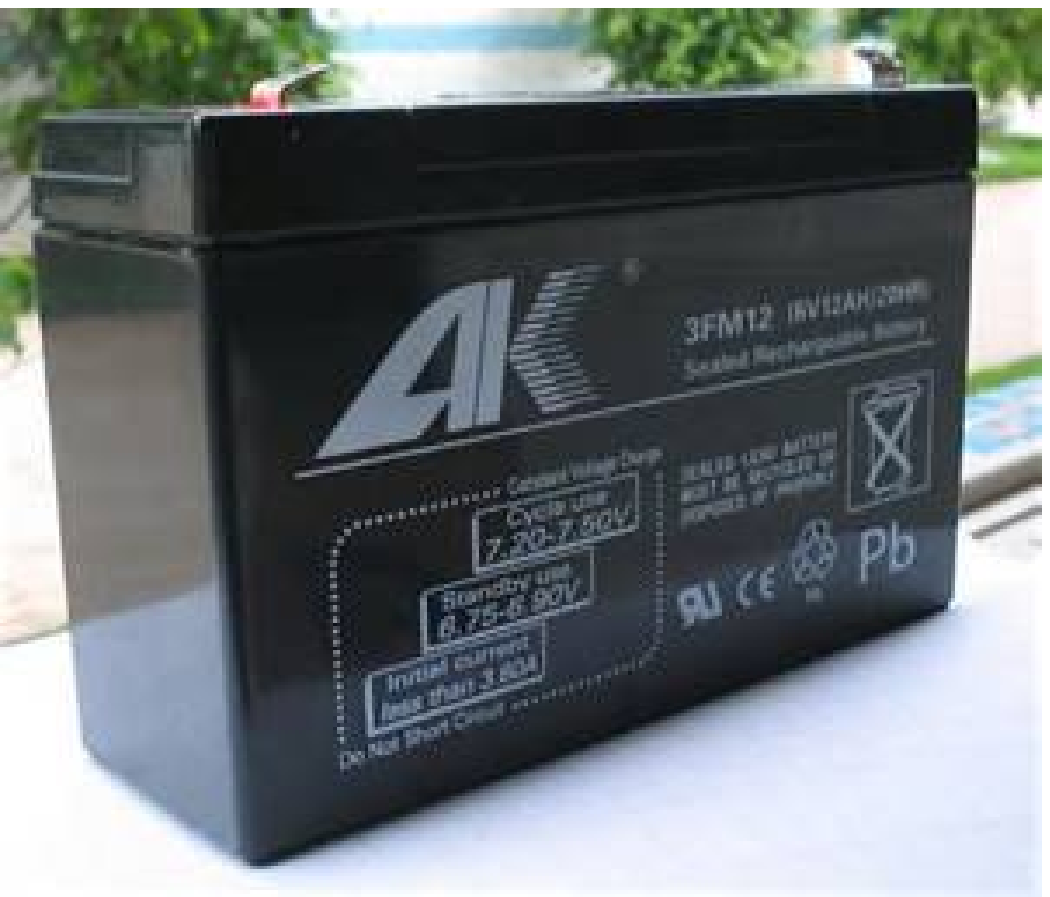}}} 
 & \multirow{3}{*}{\raisebox{-0.5\totalheight}{\includegraphics[width=1.6cm,height=1.6cm]{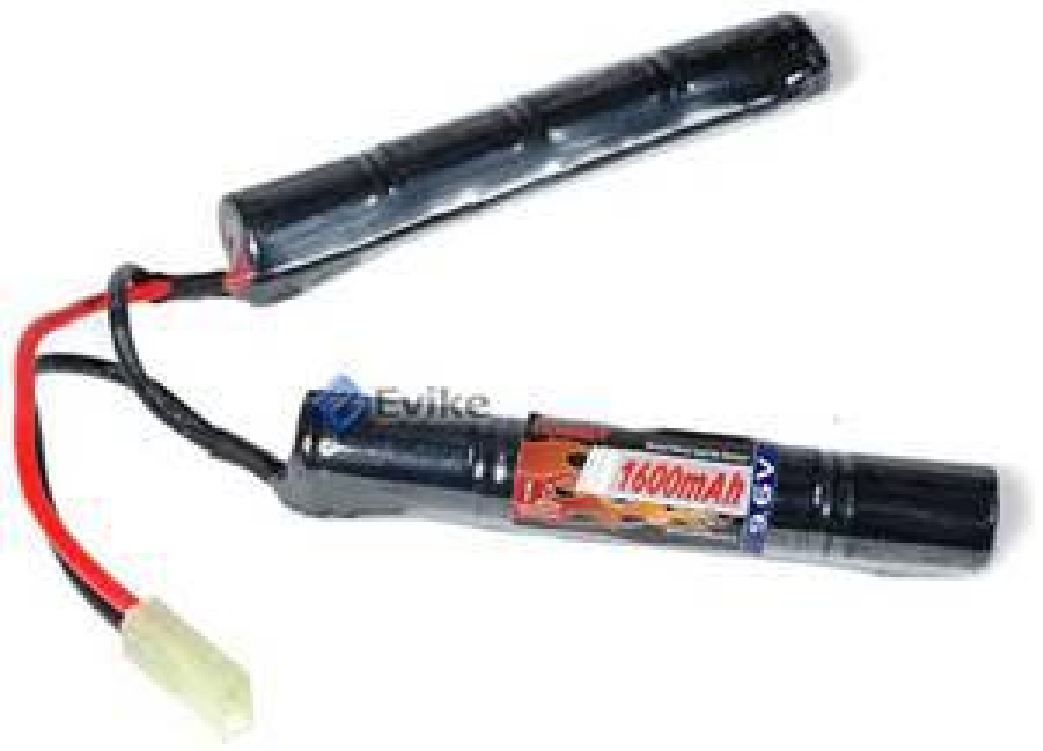}}}  \\ 
 & & &  &&&&&\\ 
 & & &   &&&&&\\ 
 & & &   &&&&&\\ 
  & & &  &$\openbigstar\openbigstar$&$\openbigstar\openbigstar$&$\bigstar\bigstar$&$\bigstar\bigstar$&$\openbigstar\openbigstar$\\
\midrule

\multirow{10}{*}{\specialcell{{[ling]} simpson \\ \\visualness: 0.5  }} & \multirow{5}{*}{text2image} & \multirow{5}{*}{0.6204} & 
\multirow{5}{*}{\specialcell{ simpson \\ ling \\ jessica simpson  \\ cody simpson \\ homer simpson }}
& \multirow{3}{*}{\raisebox{-0.5\totalheight}{\includegraphics[width=1.6cm,height=1.6cm]{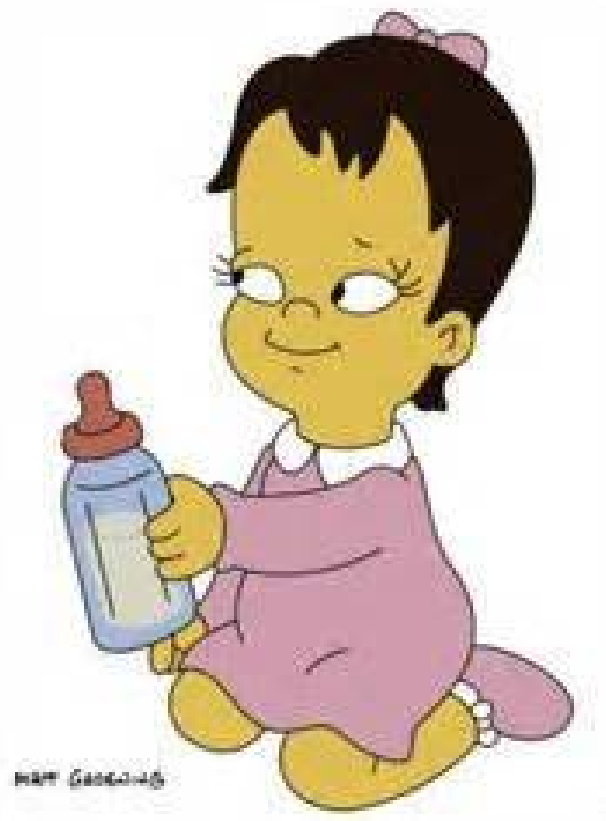}}} 
& \multirow{3}{*}{\raisebox{-0.5\totalheight}{\includegraphics[width=1.6cm,height=1.6cm]{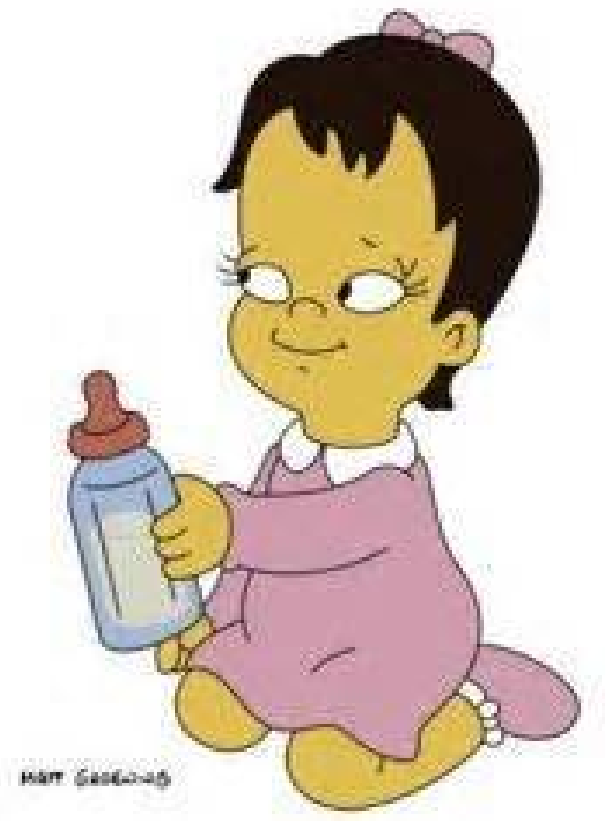}}} 
& \multirow{3}{*}{\raisebox{-0.5\totalheight}{\includegraphics[width=1.6cm,height=1.6cm]{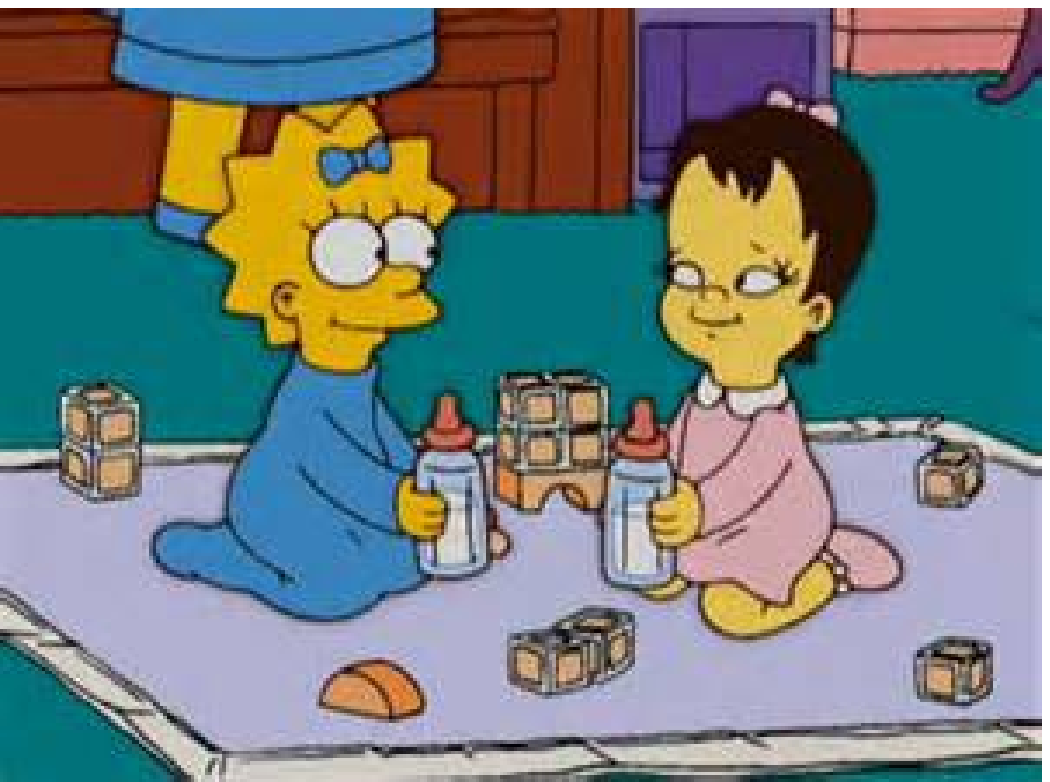}}} 
& \multirow{3}{*}{\raisebox{-0.5\totalheight}{\includegraphics[width=1.6cm,height=1.6cm]{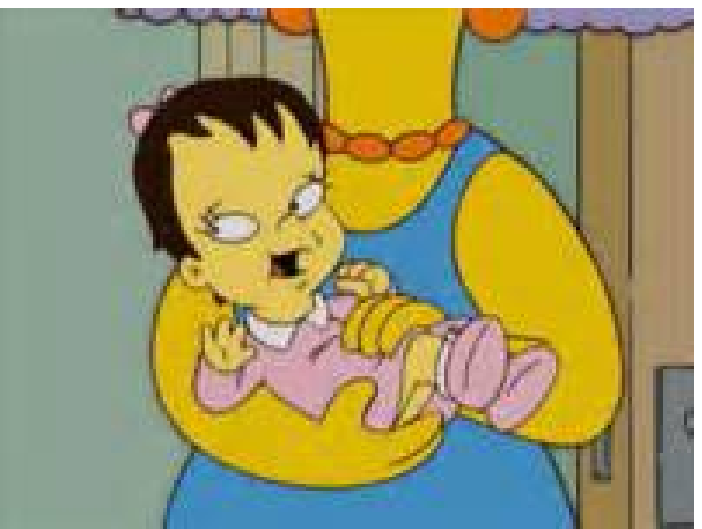}}} 
& \multirow{3}{*}{\raisebox{-0.5\totalheight}{\includegraphics[width=1.6cm,height=1.6cm]{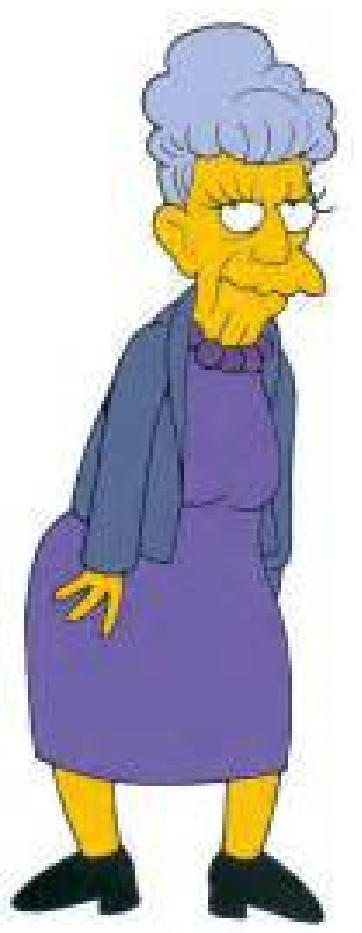}}}  \\ 
& & &   &&&&&\\ 
& & & &&&&& \\
& & &  &&&&& \\
  & & &  & $\bigstar\bigstar$& $\bigstar\bigstar$&$\bigstar\openbigstar$& $\bigstar\bigstar$&$\openbigstar\openbigstar$\\
\cmidrule(l){2-9}
 & \multirow{5}{*}{DeViSE} &  \multirow{5}{*}{0.1828} & 
 \multirow{5}{*}{\specialcell{ lisa ling \\ bridsmains ling hairstleys \\ ling \\ freshwat ling \\ ling xiaoyu }}
 & \multirow{3}{*}{\raisebox{-0.5\totalheight}{\includegraphics[width=1.6cm,height=1.6cm]{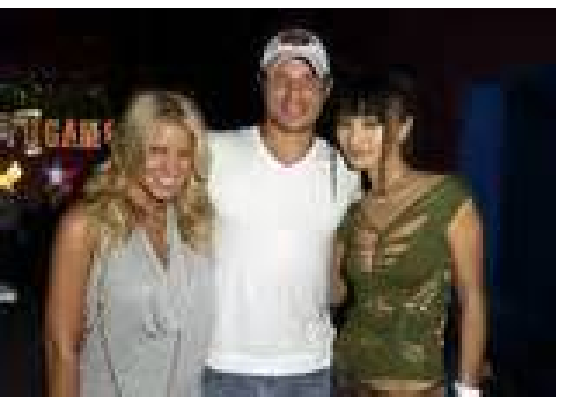}}} 
 & \multirow{3}{*}{\raisebox{-0.5\totalheight}{\includegraphics[width=1.6cm,height=1.6cm]{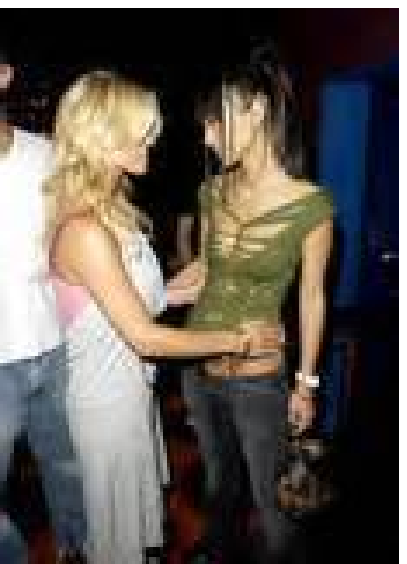}}} 
 & \multirow{3}{*}{\raisebox{-0.5\totalheight}{\includegraphics[width=1.6cm,height=1.6cm]{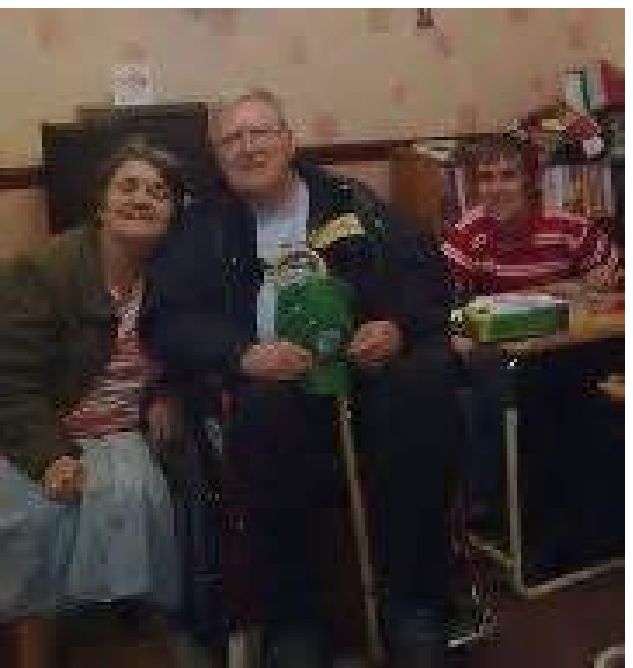}}}  
 & \multirow{3}{*}{\raisebox{-0.5\totalheight}{\includegraphics[width=1.6cm,height=1.6cm]{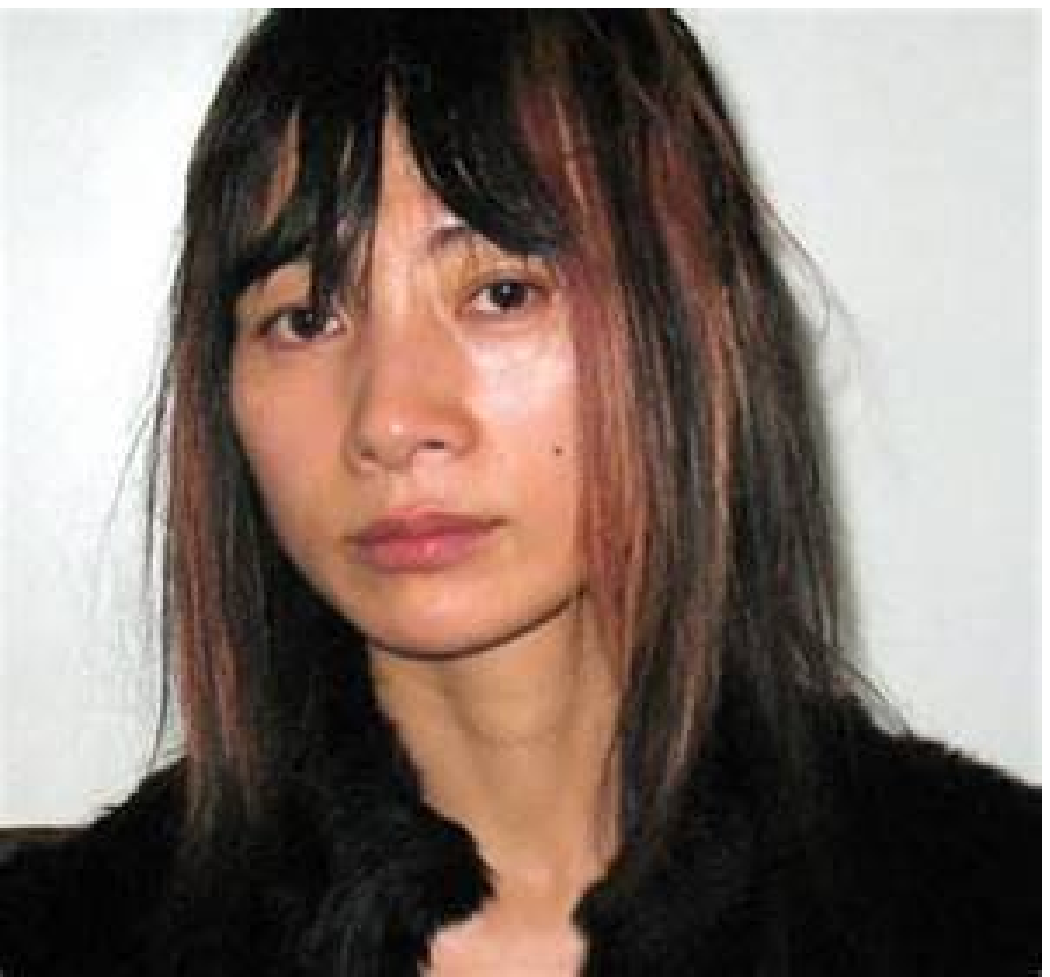}}} 
 & \multirow{3}{*}{\raisebox{-0.5\totalheight}{\includegraphics[width=1.6cm,height=1.6cm]{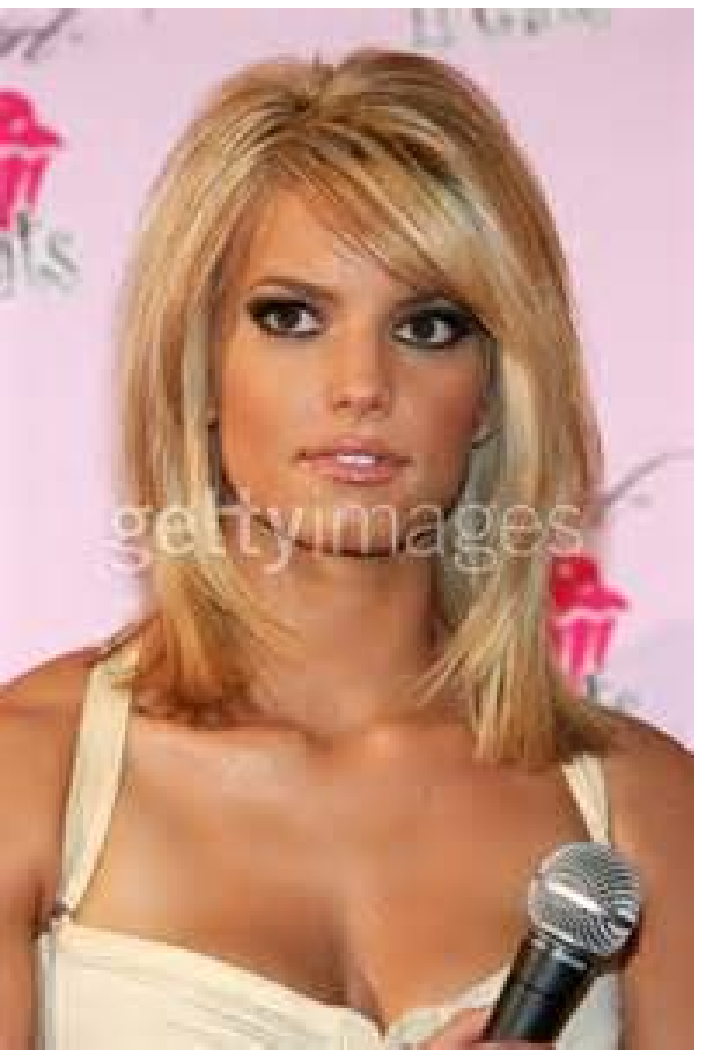}}}  \\ 
 & & &  &&&&&\\ 
 & & &  &&&&&\\
 & & &  &&&&&\\
 & & &  &$\bigstar\openbigstar$&$\openbigstar\openbigstar$&$\openbigstar\openbigstar$&$\openbigstar\openbigstar$&$\openbigstar\openbigstar$\\

\midrule


\multirow{10}{*}{\specialcell{ {[family]} \sout{photo} \\ \\visualness: 1 }} & \multirow{5}{*}{text2image} &  \multirow{5}{*}{0.2059} & 
\multirow{5}{*}{\specialcell{ family \\  family family \\ family tree \\ family guy  \\ family quote }}
& \multirow{3}{*}{\raisebox{-0.5\totalheight}{\includegraphics[width=1.6cm,height=1.6cm]{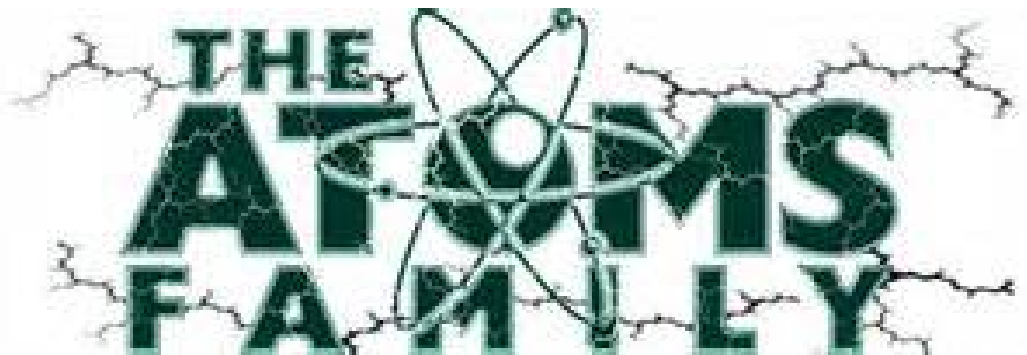}}} 
& \multirow{3}{*}{\raisebox{-0.5\totalheight}{\includegraphics[width=1.6cm,height=1.6cm]{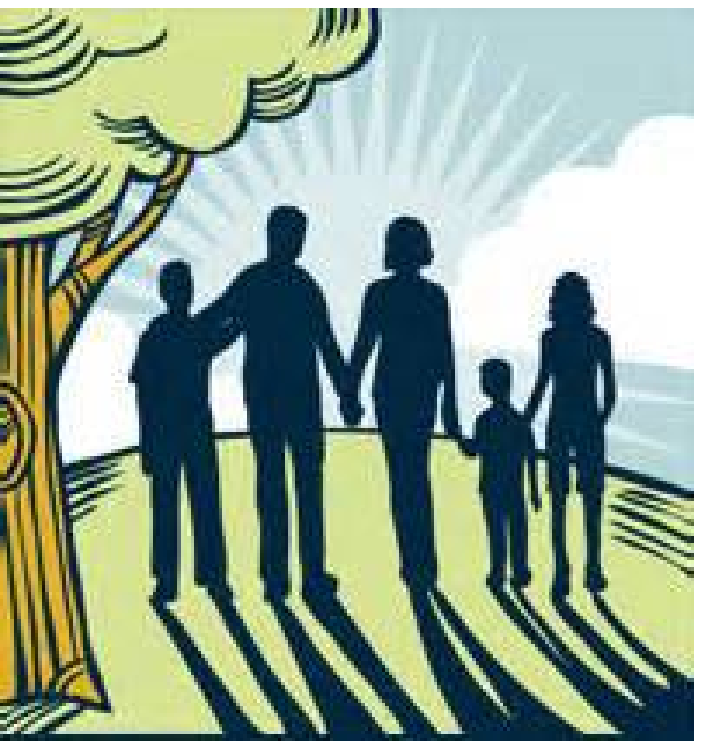}}} 
& \multirow{3}{*}{\raisebox{-0.5\totalheight}{\includegraphics[width=1.6cm,height=1.6cm]{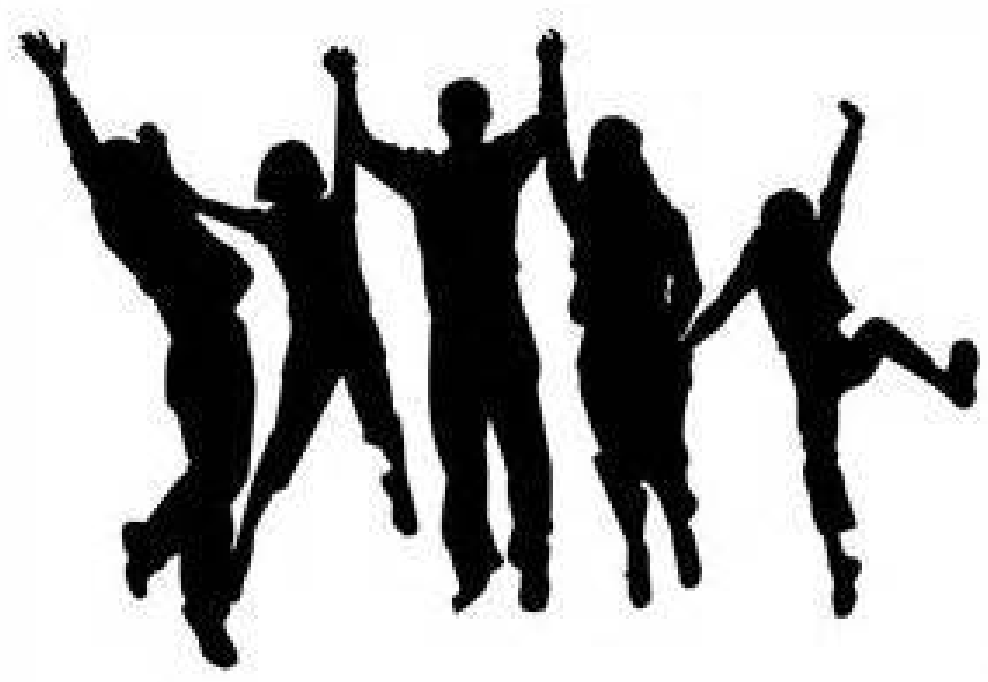}}} 
& \multirow{3}{*}{\raisebox{-0.5\totalheight}{\includegraphics[width=1.6cm,height=1.6cm]{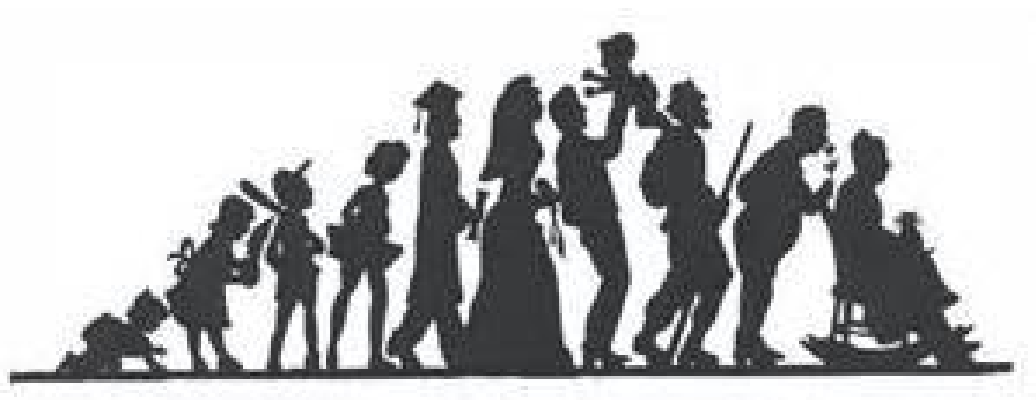}}} 
& \multirow{3}{*}{\raisebox{-0.5\totalheight}{\includegraphics[width=1.6cm,height=1.6cm]{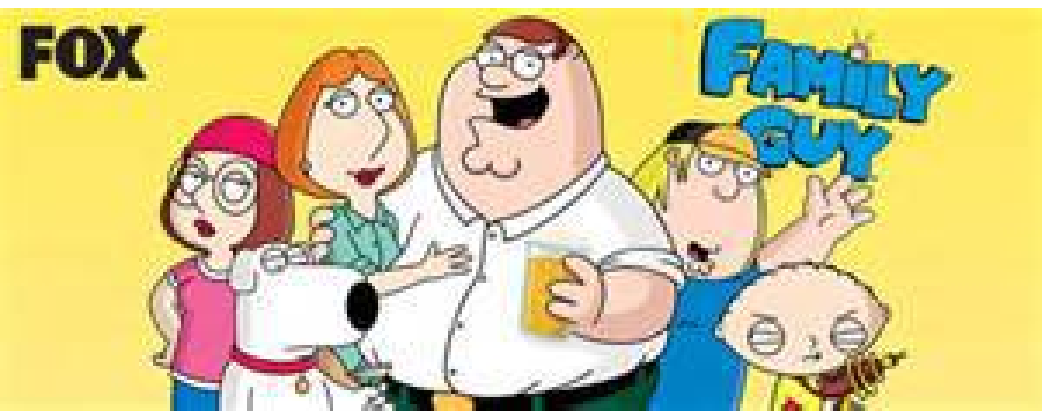}}}  \\ 
& & &  &&&&&\\ 
& & &  &&&&&\\ 
& & &  &&&&&\\ 
& & &  &$\openbigstar\openbigstar$& $\openbigstar\openbigstar$&$\openbigstar\openbigstar$& $\openbigstar\openbigstar$&$\openbigstar\openbigstar$\\
\cmidrule(l){2-9}
 & \multirow{5}{*}{DeViSE} &  \multirow{5}{*}{0.9278} & 
 \multirow{5}{*}{\specialcell{ family \\ turs family \\ tuohy family  \\ duggar family \\ santorums family}}
 & \multirow{3}{*}{\raisebox{-0.5\totalheight}{\includegraphics[width=1.6cm,height=1.6cm]{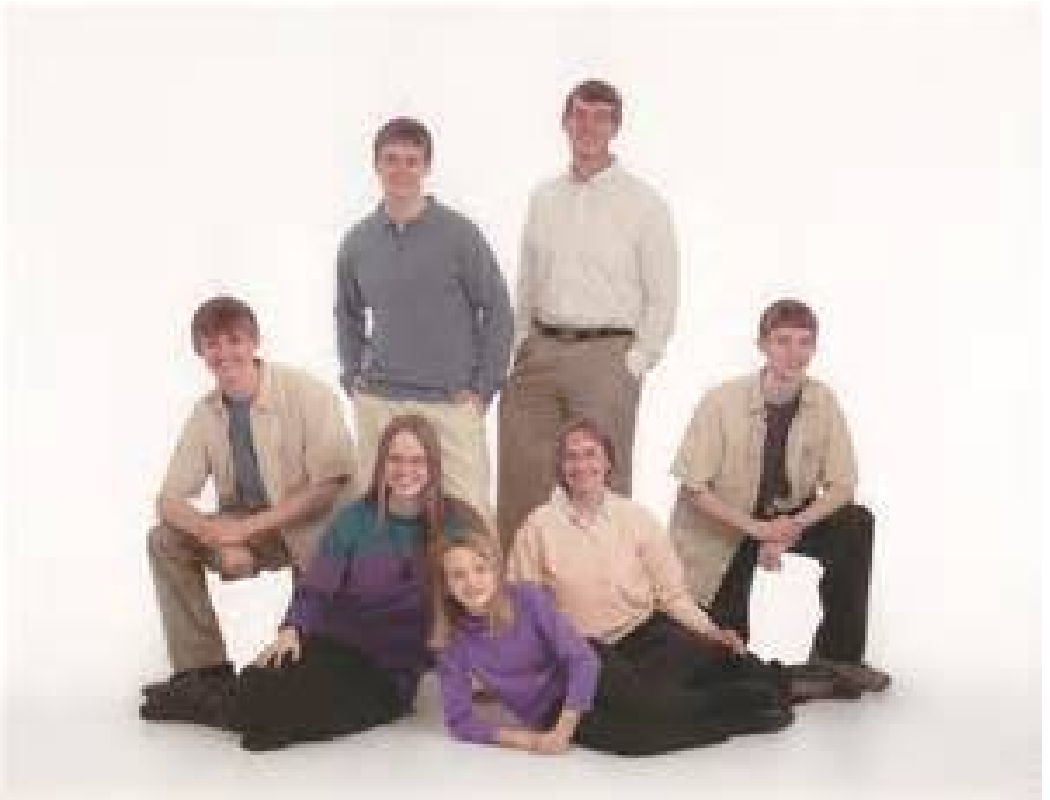}}} 
 & \multirow{3}{*}{\raisebox{-0.5\totalheight}{\includegraphics[width=1.6cm,height=1.6cm]{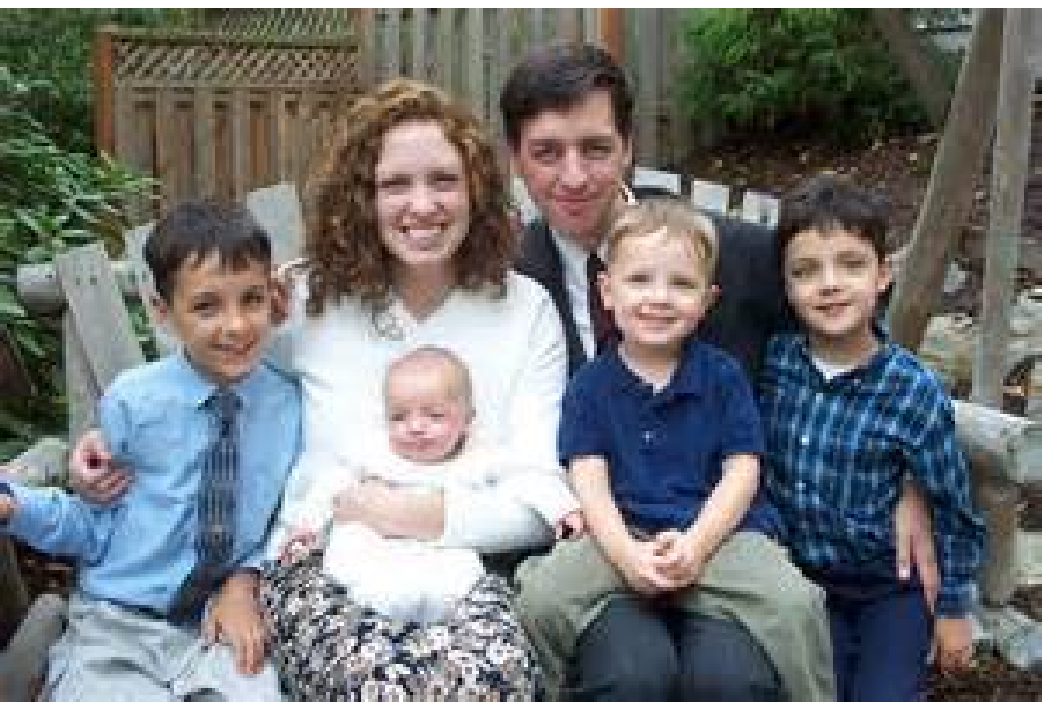}}} 
 & \multirow{3}{*}{\raisebox{-0.5\totalheight}{\includegraphics[width=1.6cm,height=1.6cm]{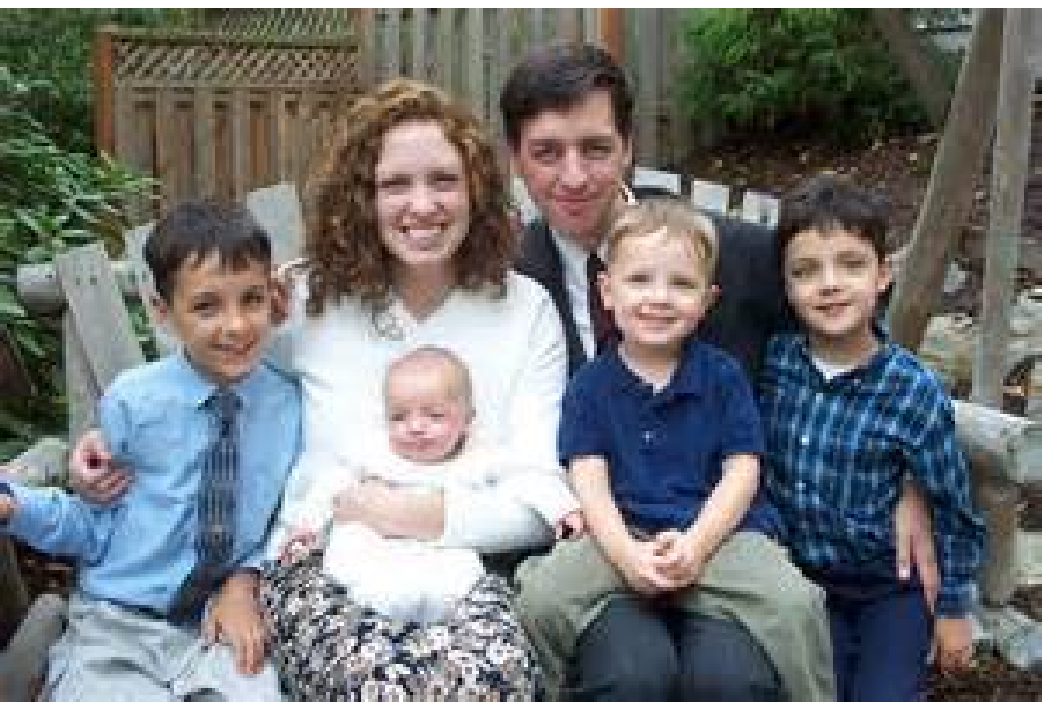}}}  
 & \multirow{3}{*}{\raisebox{-0.5\totalheight}{\includegraphics[width=1.6cm,height=1.6cm]{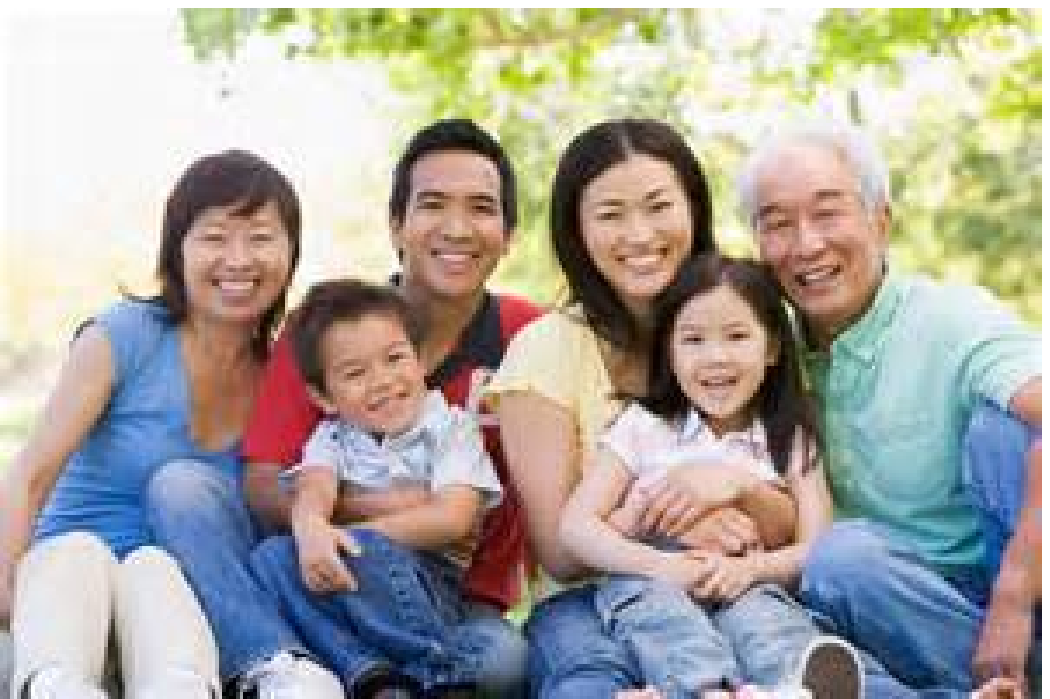}}} 
 & \multirow{3}{*}{\raisebox{-0.5\totalheight}{\includegraphics[width=1.6cm,height=1.6cm]{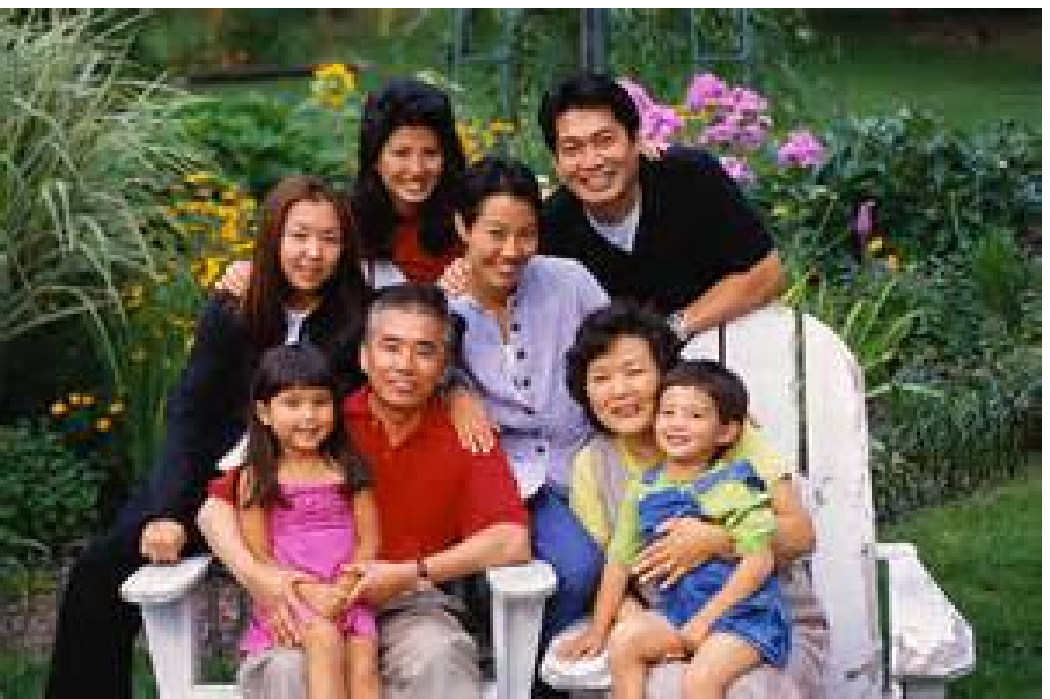}}}  \\ 
 & & &  &&&&&\\ 
 & & &  &&&&&\\ 
 & & &  &&&&&\\ 
 & & &  &$\bigstar\bigstar$&$\bigstar\bigstar$&$\bigstar\bigstar$&$\bigstar\bigstar$&$\bigstar\bigstar$\\
\midrule

\multirow{10}{*}{\specialcell{ {[woman]} {[bicycle]} \\ \\visualness: 1 }}& \multirow{5}{*}{text2image} &  \multirow{5}{*}{0.7680} & 
\multirow{5}{*}{\specialcell{ woman bicycle \\ bicycle woman \\ woman riding bicycle \\ bicycle sizing woman \\ trek woman bicycle }}
& \multirow{3}{*}{\raisebox{-0.5\totalheight}{\includegraphics[width=1.6cm,height=1.6cm]{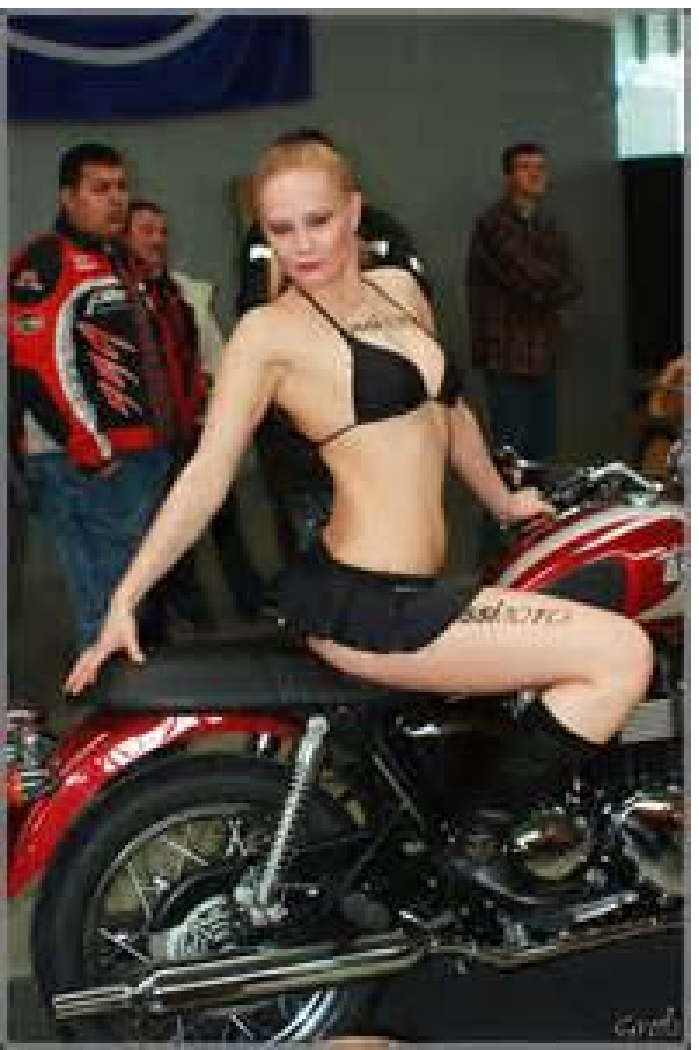}}} 
& \multirow{3}{*}{\raisebox{-0.5\totalheight}{\includegraphics[width=1.6cm,height=1.6cm]{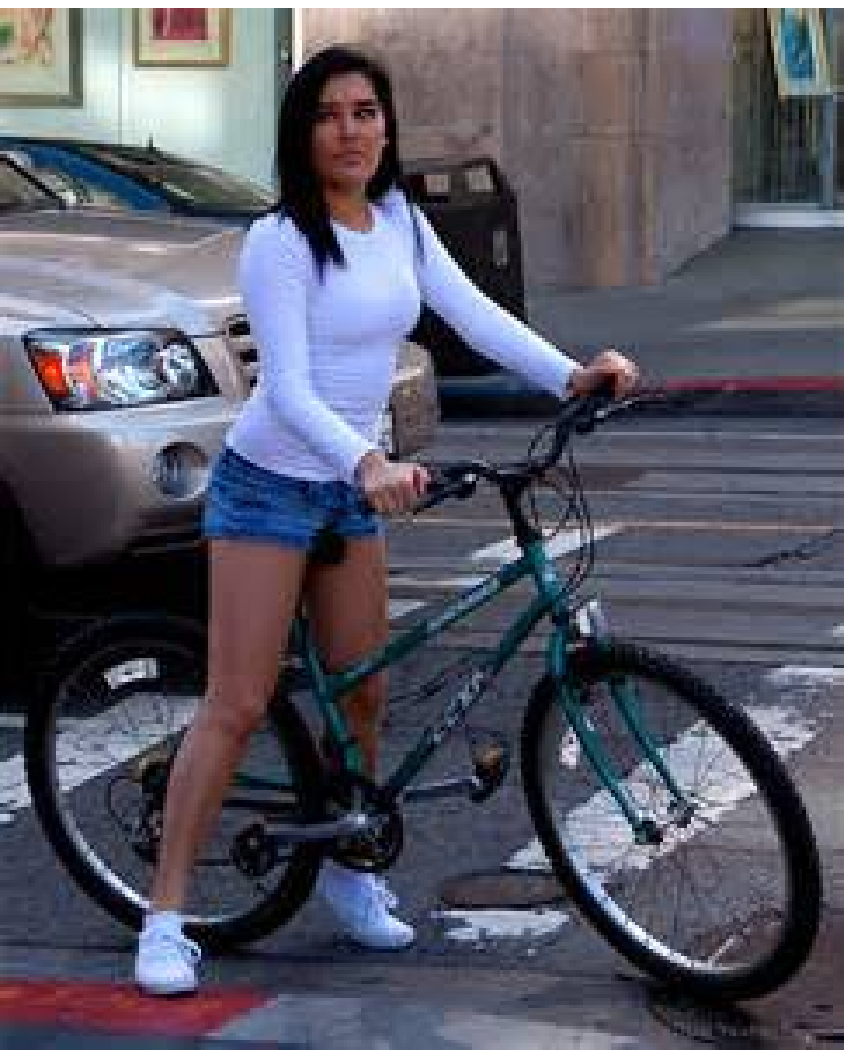}}} 
& \multirow{3}{*}{\raisebox{-0.5\totalheight}{\includegraphics[width=1.6cm,height=1.6cm]{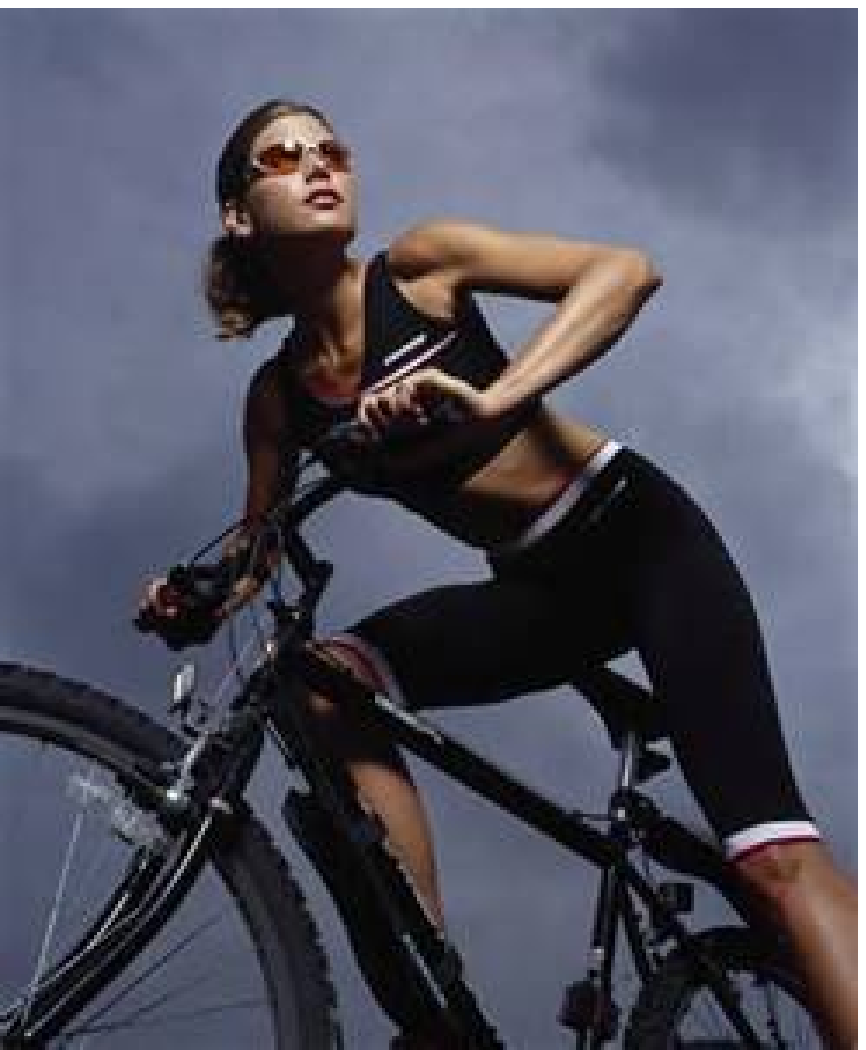}}} 
& \multirow{3}{*}{\raisebox{-0.5\totalheight}{\includegraphics[width=1.6cm,height=1.6cm]{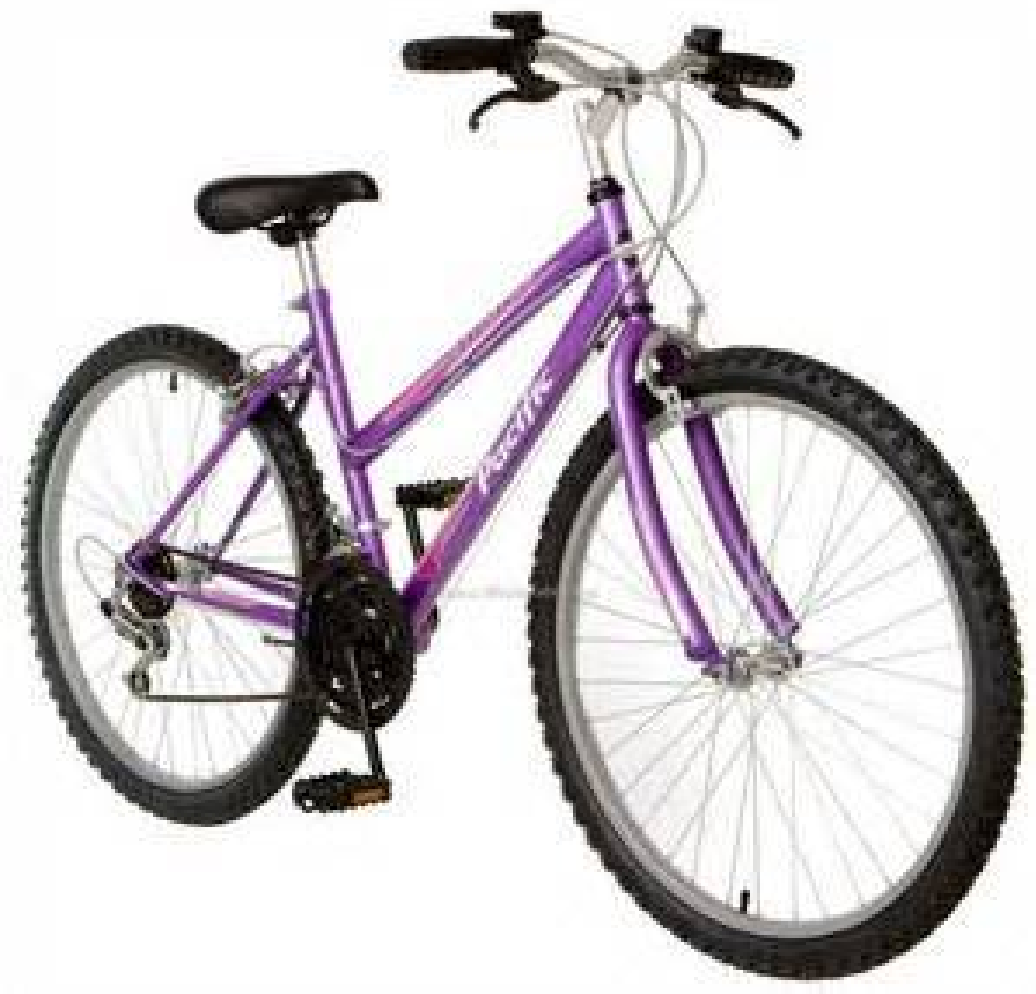}}} 
& \multirow{3}{*}{\raisebox{-0.5\totalheight}{\includegraphics[width=1.6cm,height=1.6cm]{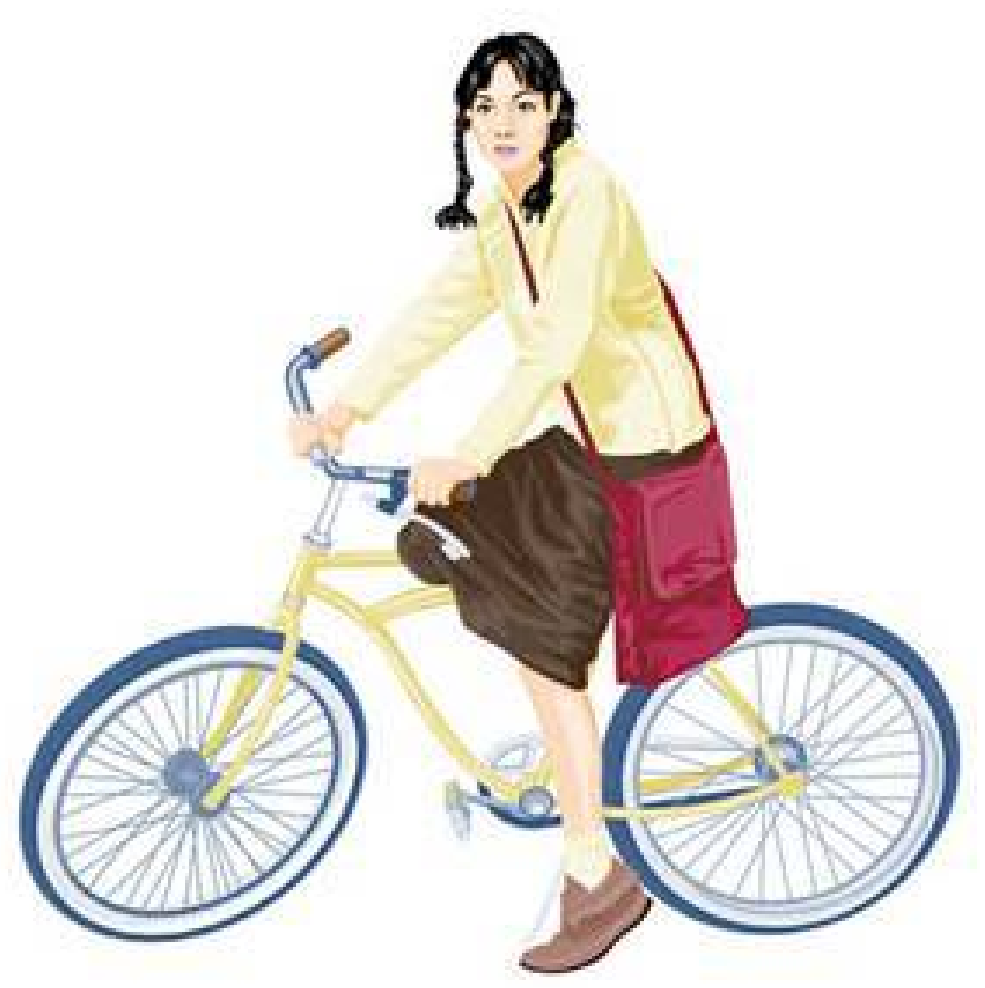}}}  \\ 
& & &  &&&&&\\ 
& & &  &&&&&\\ 
& & &  &&&&&\\
& & &  &$\openbigstar\openbigstar$& $\bigstar\bigstar$&$\bigstar\bigstar$& $\bigstar\bigstar$&$\bigstar\bigstar$\\
\cmidrule(l){2-9}
 & \multirow{5}{*}{DeViSE} &  \multirow{5}{*}{0.9169} & 
 \multirow{5}{*}{\specialcell{ woman bicycle \\ bicycle woman \\ bike woman  \\ woman bike \\ bike woman bewach crusiers }}
 & \multirow{3}{*}{\raisebox{-0.5\totalheight}{\includegraphics[width=1.6cm,height=1.6cm]{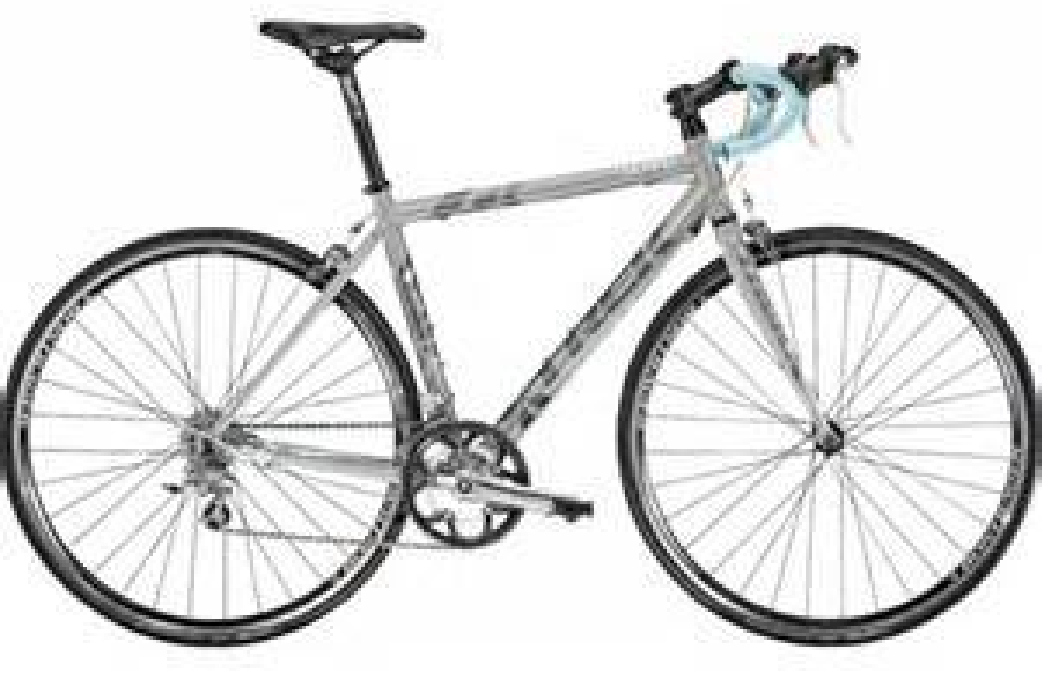}}} 
 & \multirow{3}{*}{\raisebox{-0.5\totalheight}{\includegraphics[width=1.6cm,height=1.6cm]{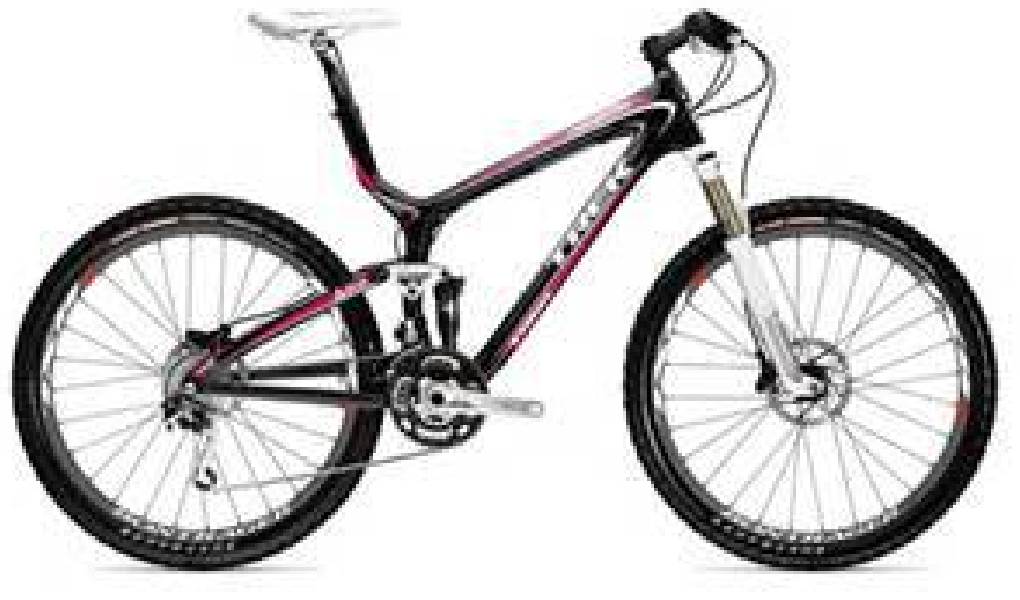}}} 
 & \multirow{3}{*}{\raisebox{-0.5\totalheight}{\includegraphics[width=1.6cm,height=1.6cm]{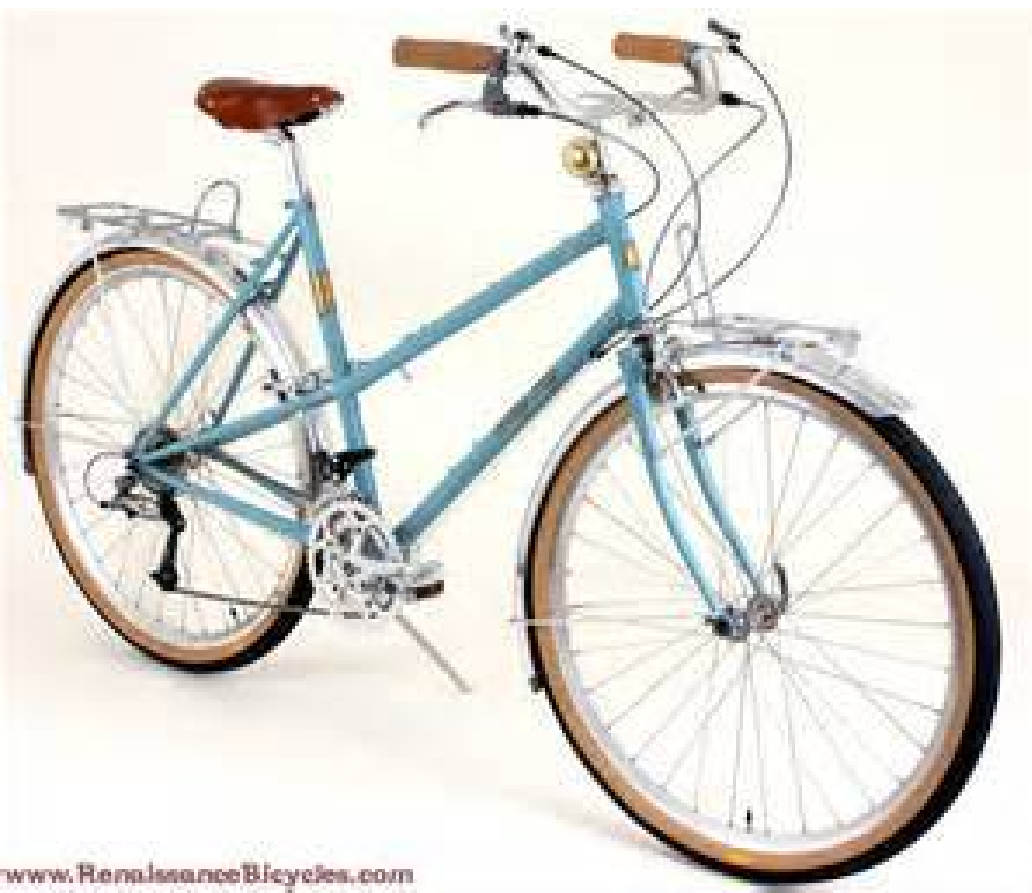}}}  
 & \multirow{3}{*}{\raisebox{-0.5\totalheight}{\includegraphics[width=1.6cm,height=1.6cm]{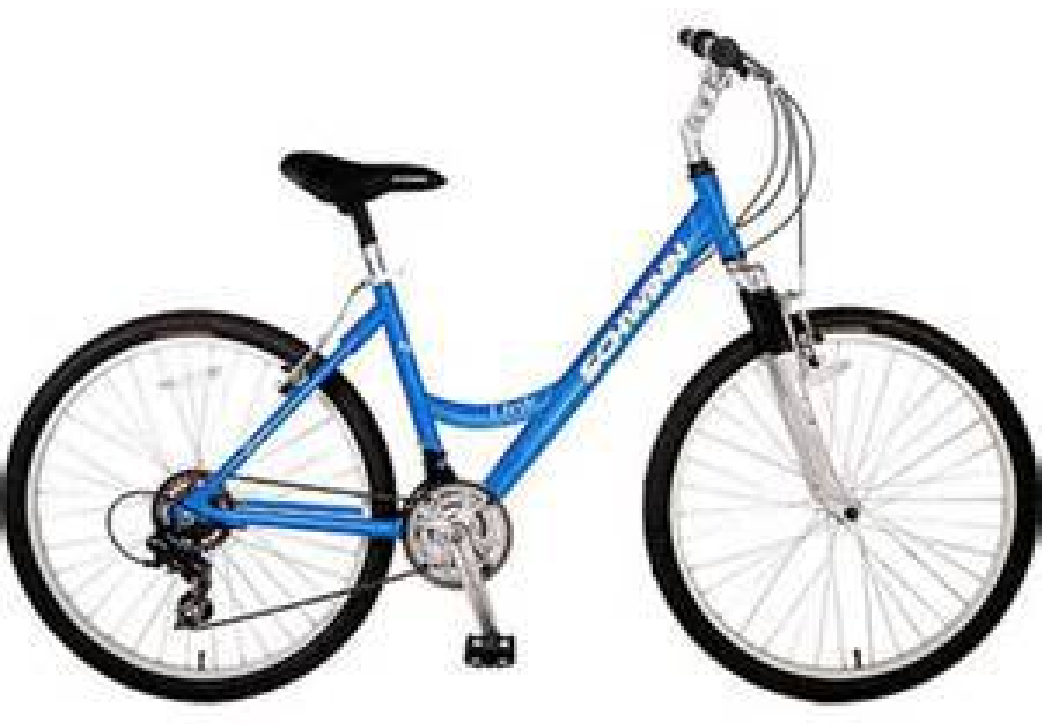}}} 
 & \multirow{3}{*}{\raisebox{-0.5\totalheight}{\includegraphics[width=1.6cm,height=1.6cm]{search_example/34966}}}  \\ 
 & & &  &&&&&\\ 
 & & &  &&&&&\\ 
 & & &  &&&&&\\ 
 & & &  &$\bigstar\bigstar$&$\bigstar\bigstar$&$\bigstar\bigstar$&$\bigstar\bigstar$&$\bigstar\bigstar$\\


\bottomrule
\end{tabular}
}
\end{table*}

\textbf{The influence of fusion}. Similar to Experiment 1, we investigate method-fusion and feature-fusion.
As shown in Table \ref{tab:irc_ndcg}, in general fusion gains some performance improvement.
Nonetheless, the difference between the single best run (text2image + CaffeNet, MAP of 0.5153),
and average-fusion (MAP of 0.5177) does not pass the significance test.
This again differs from the results of Experiment 1, where fusion brings in clear improvement.

\begin{figure}[tb!]
\centering\includegraphics[width=0.9\columnwidth]{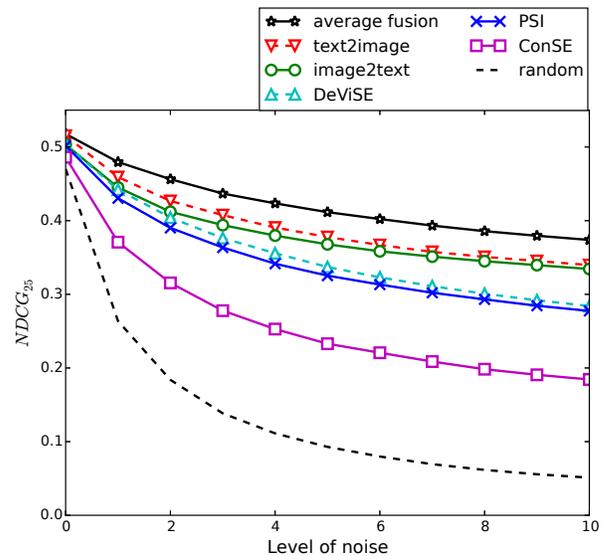}
\caption{\textbf{Performance curves of different methods with respect to the artificial noise}. Test set: Clickture-dev. The two baseline methods outperform the three advanced methods given a specific level of noise.}\label{fig:robust_analysis}
\end{figure}

\textbf{Robust analysis}.
As already shown in Fig. \ref{fig:pos_neg_3}, for most queries more than half of the test images are Excellent or Good,
meaning a random sort may return relevant test images with a good chance.
We analyze the robustness of each method by adding extra noise.
Concretely, for each query with $n$ test images,
we add $h$-fold noise, namely $h \times n$ images randomly taken from other queries,
with $h=1,2,\ldots,10$. The performance curves with respect to the level of noise are shown in Fig. \ref{fig:robust_analysis}.
Unsurprisingly, for all methods the performance goes down. Yet, the curve of the random run drops more sharply than the others,
indicating a more challenging start point. Given the varied levels of noise,  our conclusion that the baselines are better than the advanced models still holds. Interestingly, the influence of fusion is more evident now.
The result suggests that fusion robustifies cross-media similarity computation.

\begin{figure}[tb!]
\centering\includegraphics[width=0.9\columnwidth]{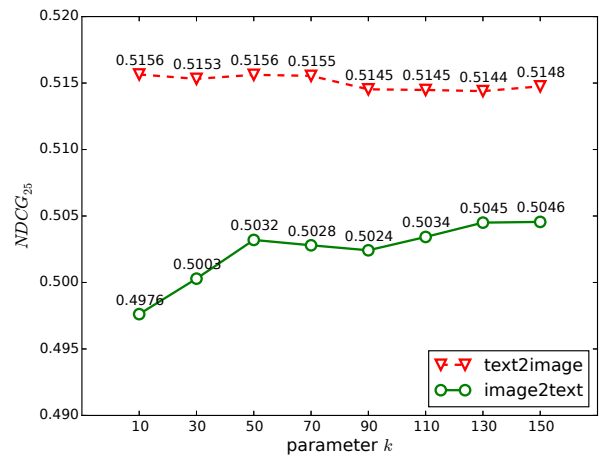}
\caption{\textbf{The influence of the parameter $k$ on the two baseline methods},
showing that image2text benefits more from optimizing $k$. Test set: Clickture-dev. }\label{fig:k_effect}
\end{figure}

Previously we have empirically set the parameter $k$, which corresponds to the number of neighbor images for image2text and the number of neighbor queries for text2image, to be 50 and 30, respectively. To reveal the influence of this parameter on the two baselines, we try $k$ with its value ranging from 10 to 150 with an interval of 20. 
As shown in Fig. \ref{fig:k_effect}, image2text benefits more from optimizing $k$.

\textbf{Efficiency analysis}.
We implement all the five models in python, and employ the theano deep learning library for PSI and DeViSE.
Additionally, for image2text we employ the production quantization algorithm \cite{jegou2011product} to accelerate $k$ visual neighbor search.
Clickture-Lite is used as the training data and Clickture-dev as the test data.
As visual feature extraction is required by all the methods and can be precomputed, we exclude this part from the comparison.
Table \ref{tab:train-test-time} provides computational cost and memory requirements of these methods on a regular PC with 32G RAM and a GTX TITANX X GPU. The trained DeViSE model is more compact and faster than text2image.

\begin{table} [tb!]
\renewcommand{\arraystretch}{1.2}
\caption{\textbf{Computational cost and memory requirements of the five models}.  Training set: Clickture-Lite. Test set: Clicture-dev. Visual feature: CaffeNet. ConSE requires no training and is the most efficient for cross-media similarity computation.}
\label{tab:train-test-time}
\centering \scalebox{0.9}{
\begin{tabular}{@{}ll r rr r r@{}}
\toprule
&&  \multicolumn{2}{c}{Training}  &&   \multicolumn{2}{c}{Test} \\
\cmidrule{3-4}  \cmidrule{6-7}
Method          &&  Time     &  Memory   &&  Time   &  Memory    \\
\cmidrule{1-1} \cmidrule{3-7} 
image2text      && -         & -         &&  4.4 hours   & 3,800 M  \\
text2image      && -         & -         &&  980 seconds & 4,500 M  \\
PSI             && 120 hours & 1450 M    &&  550 seconds & 750 M  \\
DeViSE          && 50 hours  & 1400 M    &&  385 seconds & 670 M  \\
ConSE           && -         & -         &&  110 seconds & 400 M  \\
\bottomrule
\end{tabular}
}
\end{table}

\textbf{Performance on IRC-MM15-test}.
As aforementioned, the possibility of overfitting on Clickture-dev exists because of its full availability.
Hence, we further evaluate on IRC-MM15-test, the ground truth of which is unavailable to us.
Table \ref{tab:dcg_test} presents the performance of the selected methods.
The weights of learned-fusion are optimized on Clickture-dev.
The result again confirms our finding that text2image surpasses DeViSE for answering real-user queries.
The difference between the two is statistically significant.
In addition, the best run also outperforms our conference version \cite{dong2015fusion} in post-competition evaluation\footnote{The best run of  \cite{dong2015fusion} obtains {NDCG}\textsubscript{25} of 0.5200, using a search result re-ranking trick. 
Adding the same trick to our solution has scored NDCG\textsubscript{25} of 0.5312. However, this trick does not work when given individual image-query pairs, so we exclude it from the comparison.}.

\begin{table} [tb!]
\renewcommand{\arraystretch}{1.2}
\caption{\textbf{Performance on IRC-MM15-test}.}
\label{tab:dcg_test}
\centering \scalebox{1.0}{
\begin{tabular}{@{}ll  l@{}}
\toprule
%
Method  &&  {NDCG}\textsubscript{25}\\
\cmidrule{1-1} \cmidrule{3-3} 
Random baseline              && 0.4260 \\
Upper bound                  && 0.6924 \\ 
Top performer \cite{dong2015fusion}  && 0.4929 \\ [3pt]

\emph{This work:} \\
DeViSE           && 0.4842  \\
text2image         && 0.4902  \\
average-fusion               && 0.4946 \\
learned-fusion               && \textbf{0.4963} \\
\bottomrule
\end{tabular}
}
\end{table}

\subsection{Experiment 3. Analytics using Query Visualness} \label{ssec:exp-query-analytics}

\begin{figure*}[tb!]
\centering
 \subfigure [Clickture-dev] {
\noindent\includegraphics[width=0.85\columnwidth]{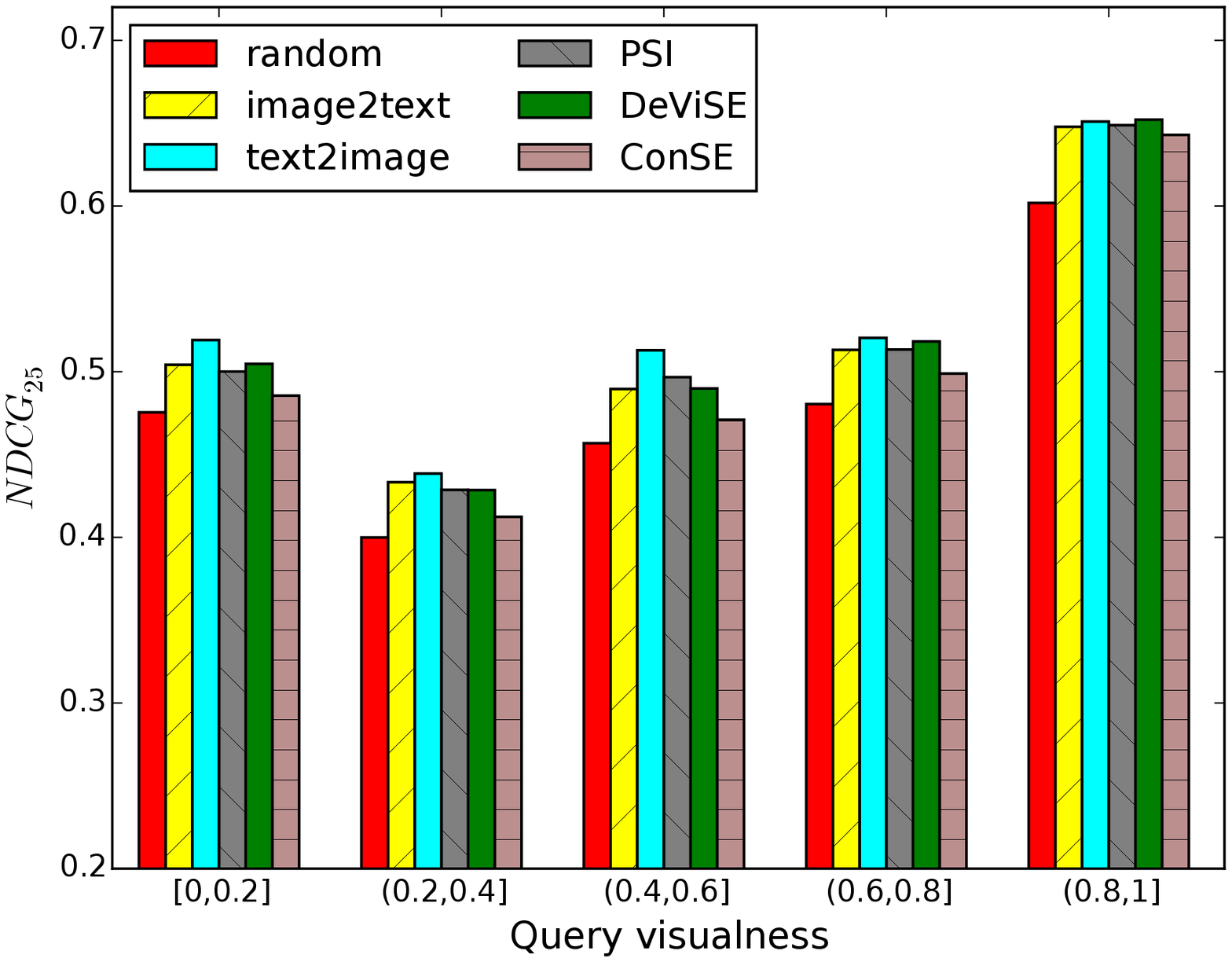}
\label{fig:msr2013dev_visual_perf}}
 \subfigure[Adding one-fold noise to Clickture-dev] {
\noindent\includegraphics[width=0.85\columnwidth]{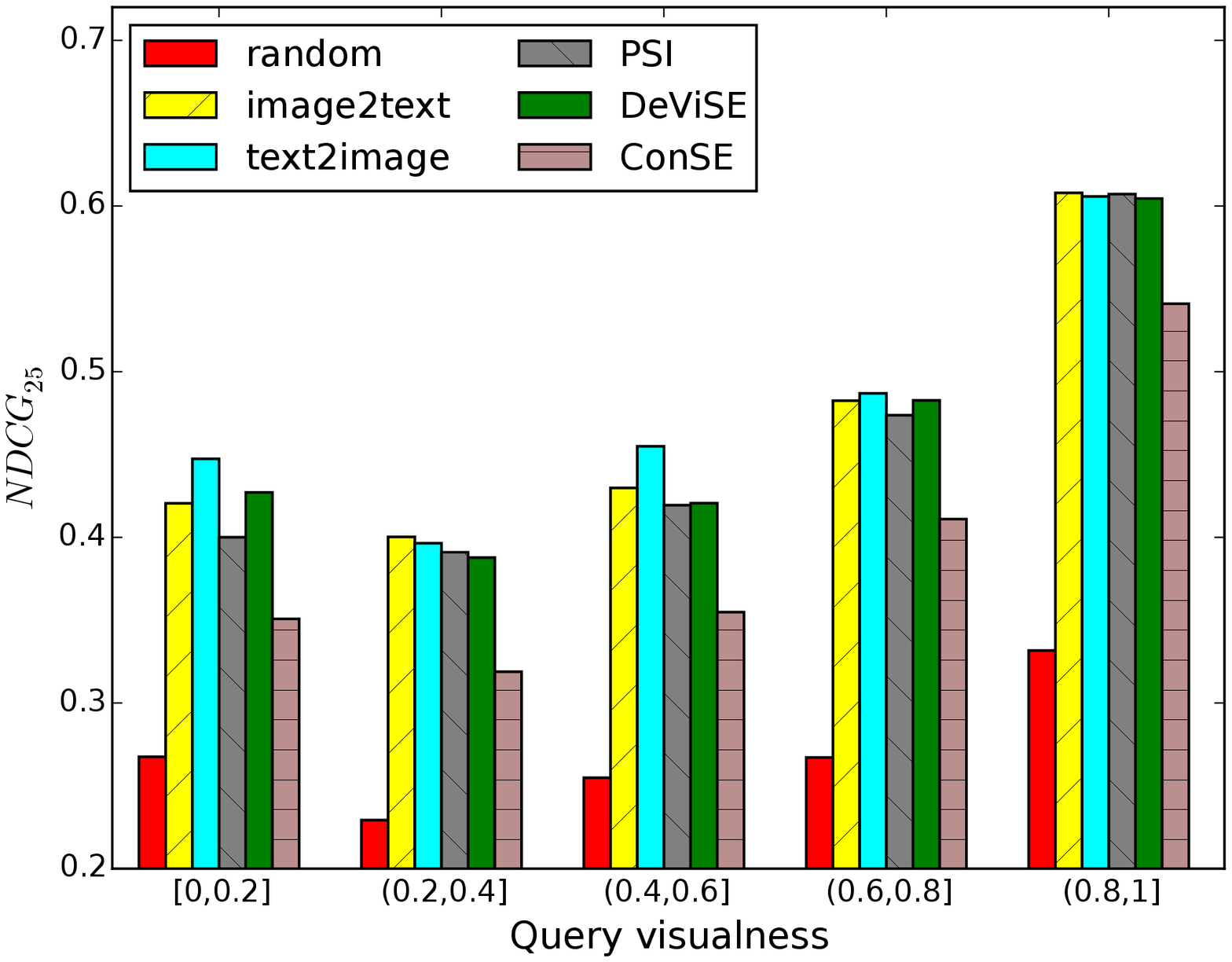}
\label{fig:msr2013noisy1_visual_perf}}
\caption{
\textbf{Performance of different models on (a) Clickture-dev and (b) Clickture-dev with one-fold noise}.
Queries are grouped according to their visualness. The minimal performance is for query visualness in (0.2, 0.4) due to the fact that this group has the lowest percentage of relevant images, see Fig. \ref{fig:pos_neg_3}, and thus image retrieval for this group is more challenging. Observing (b) from left to right, the average gain of the five models over the random run increases along with query visualness, from 0.1416, 0.1495, 0.1611 to 0.2003 and 0.2616, showing the current models better address visual-oriented queries.}
 \label{fig:robust_analyse}
\end{figure*}

Thus far all the comparisons are holistic. 
To gain a further understanding of the individual methods,
we leverage query visualness developed in Section \ref{ssec:query-log}.
The 1,000 test queries from Clickture-dev are grouped according to their visualness, 
with the performance of each group shown in  Fig. \ref{fig:msr2013dev_visual_perf}.
DeViSE outperforms the two baselines for queries with visualness scores over 0.8.
The result is in line with what we have observed in Experiment 1.

Still, Fig. \ref{fig:msr2013dev_visual_perf} does not allow us to conclude if visual-oriented queries are better handled, 
because the random run already gives relatively high NDCG of 0.6118.
So we add one-fold noise to make the random runs more balanced across different groups.
Note that queries having more relevant images in the test set have higher AP scores a priori. So the performance of the random run of the five groups need to be taken into account. 
Observing Fig. \ref{fig:msr2013noisy1_visual_perf} from left to right, the average gain of the five models over the random run increases along with query visualness, from 0.1416, 0.1495, 0.1611 to 0.2003 and 0.2616.
The gain on the group of the lowest visualness is the smallest, while the group of the highest visualness is the largest. Hence, the current models better address visual-oriented queries.
%
Moreover, since the notion of query visualness is orthogonal to the development of the cross-media models, its connection to the model performance indicates that such a query categorization is meaningful.

\begin{figure}[tb!]
\centering\includegraphics[width=0.9\columnwidth]{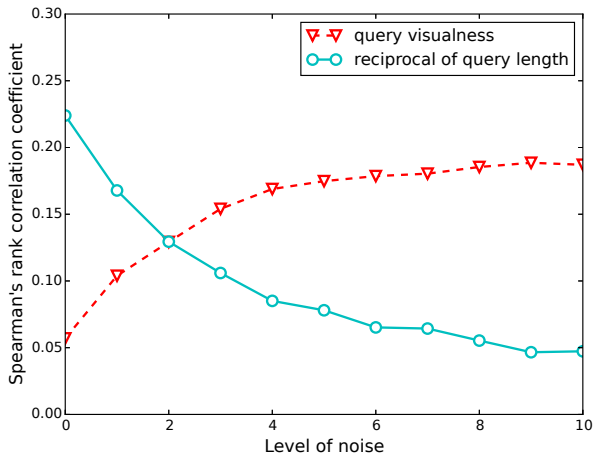}
\caption{\textbf{Spearman's rank correlation coefficients between query visualness / reciprocal of query length and the performance of DeViSE, under different levels of noise}. Test set: Clickture-dev. As the task becomes more difficult, query visualness exhibits stronger monotonic correlation to the performance.}\label{fig:corre}
\end{figure}

To further verify the necessity of the proposed query visualness measure, we check if a similar connection can be found with query length, a property frequently discussed in query log analysis. To that end, we employ Spearmman's rank correlation, which provides a nonparametric measure of the monotonic relationship between two ranked variables. The correlation is computed by accumulating squared difference in paired ranks, so a perfect correlation of +1 or −1 occurs if each of the variables is a perfect monotone function of the other. 
In our context, the variables are the 1,000 test queries from Clickture-dev, with their ranks obtained by sorting in terms of three criteria separately, \ie the performance of DeViSE, query visualness, and reciprocal of query length. Note that query length tends to be negatively correlated with the performance, so we use its reciprocal for the ease of comparison. Consequently, we compute the Spearmman correlation between the first and the second ranked lists and between the first and the third ranked lists. As shown in Fig. \ref{fig:corre}, given zero noise the reciprocal of query length, with the coefficient of 0.224, appears to be better correlated to the performance when compared to query visualness with the coefficient of 0.057. Looking into Clickture-dev, we find that the test images of shorter queries, in particular, with one or two words,  contain many more Excellent examples.  So shorter queries have better performance a priori. However, as more noise is added, the influence of such a bias is lessened. This explains why the reciprocal of query length has larger correlations at the beginning, but is surpassed by query visualness later. As the task becomes more difficult, query visualness exhibits larger monotonic correlation to the performance.

As we have mentioned in Section \ref{ssec:query-log},
celebrity-related queries receive low visualness scores due to the limit of our visual concept vocabulary.
Inspired by \cite{wu2013search}, we identify these queries in Clickture-dev by a semi-automatic approach as follows.
Using a list of 2,657 celebrities from the Internet\footnote{\url{http://www.posh24.com/celebrities/a_to_z}},
we first build a name vocabulary by putting their first and last names together.
Accordingly, we obtain over 400 test queries having at least one word from the vocabulary.
A list of 240 celerity-related queries is compiled after manual verification.
Performance of these queries is given in Table \ref{tab:cele_dcg},
where text2image again performs the best.

\begin{table} [tb!]
\renewcommand{\arraystretch}{1.2}
\caption{\textbf{Performance of celebrity-related queries on Clickture-dev}.
Methods sorted in terms of their {NDCG}\textsubscript{25} scores.}
\label{tab:cele_dcg}
\centering \scalebox{1.0}{
\begin{tabular}{@{}ll  rr@{}}
\toprule
Method && NDCG\textsubscript{25}  \\
\cmidrule{1-1}  \cmidrule{3-3} 
random   &&  0.4423   \\ [3pt]
ConSE             &&  0.4441   \\
image2text        &&  0.4638   \\
DeViSE            &&  0.4611   \\
PSI               &&  0.4655   \\
text2image        &&  \textbf{0.4944}   \\
\bottomrule
\end{tabular}
}
\end{table}

\subsection{Experiment 4. Comparison to the State-of-the-Art} \label{ssec:exp-versus-others}

Given that we have evaluated only three semantic embedding models,
it would be bold to claim text2image as a new baseline.
So in this part,  we compare with a number of state-of-the-art works that report their performance on Clickture-dev,
and thus the numbers are directly comparable. The works are: \\
\emph{1) CCA} \cite{pan2014click}: Find the transformation matrices that maximizes the correlation between embedding vectors of relevant image-query pairs. \\
\emph{2) CCL} \cite{pan2014click2}: Learn a latent space by minimizing the distance of relevant image-query pairs in the new space,
while preserving the structure in the original feature space. \\
\emph{3) CSM} \cite{yu2015learning}: Project images and queries into a latent space by a deep CNN and an affine transformation, respectively. \\
\emph{4) BoWDNN} \cite{bai2014bag}: Extract a bag-of-words vector from an input image using a deep CNN. \\
\emph{5) MRW-NN} \cite{wu2016learning}:  A graph-based representation learning algorithm to generate a common space 
wherein images and queries strongly connected in click-through logs are close.  \\ 
\emph{6) RCCA} \cite{iccv2015-yaolearning}: An improved version of CCA, first learning a common space by CCA and then adjusting the space to preserve preference relationships in click-through data. \\
%



The performance of the above works is listed in Table \ref{tab:dcg_dev}. 
Our methods produce larger NDCG\textsubscript{25} scores.
Due to the lack of per-query scores, we are unable to conclude if the difference is significant.
Nevertheless, we empirically find out that when the performance difference between two image retrieval systems, \ie the $diff$ to be compared in the randomization test, is larger than 0.005, it is often sufficient to pass the significance test.
Hence, our fusion results are likely to be significantly better than the state-of-the-art.

\begin{table} [tb!]
\renewcommand{\arraystretch}{1.2}
\caption{\textbf{Comparing with the-state-of-art on Clickture-dev}. The proposed text2image is on par with the state-of-the-art, and can be further improved by simple average-fusion.}
\label{tab:dcg_dev}
\centering \scalebox{1.0}{
\begin{tabular}{@{}ll  l@{}}
\toprule
%
Method && {NDCG}\textsubscript{25} \\
\cmidrule{1-1} \cmidrule{3-3} 
CCA         \cite{pan2014click}     && 0.5055    \\ 
CCL         \cite{pan2014click2}    && 0.5059    \\ 
CSM         \cite{yu2015learning}   && 0.5070    \\ 
BoWDNN      \cite{bai2014bag}       && 0.5089    \\               
MRW-NN  \cite{wu2016learning}       && 0.5104   \\ 
RCCA        \cite{iccv2015-yaolearning}      && 0.5112    \\    [3pt]         

\emph{This work:} \\
text2image        && 0.5153   \\
average-fusion    && \textbf{0.5177}   \\
\bottomrule
\end{tabular}
}
\end{table}

In addition, we compare the proposed text2image method with two alternatives, namely query-based scoring \cite{fang2013towards} which uses all images from the neighbor queries and online classification \cite{wang2014france} using only the most similar images to a test image.
Given the same CaffeNet feature, our method with {NDCG}\textsubscript{25} of 0.5153 is significantly better than the two alternatives which score {NDCG}\textsubscript{25} of 0.4905 and 0.4958, respectively.

\section{Summary and Conclusions} \label{sec:conclusion}

As an initial effort to quantify progress on web image retrieval, 
this paper presents a systematic study that combines large-scale query log analysis and an  state-of-the-art evaluation of cross-media similarity models. 
The proposed query visualness measure helps answering \textit{Q1} in two aspects, \ie as a computable property to categorize large-scale queries and a better indicator of image retrieval performance than query length.
Given the proposed text2image method as the baseline,
much progress has already been made by the advanced semantic embedding models.
Nonetheless, the progress is mainly attributed to their relatively good performance on visual-oriented queries (\textit{Q2}). This class of queries accounts for only a small part of real-user queries.
Image retrieval experiments on the Clickture dataset show that text2image outperforms several recently developed deep learning models including DeViSE, ConSE, BoWDNN, MRW-NN, and RCCA.
For web image retrieval in the wild, we recommend text2image as a new baseline to be compared against when one advocates novel cross-media similarity models.


\section*{Acknowledgment}
The authors are grateful to the MSR-IRC organizers (Dr. Lei Zhang and Dr. Yuxiao Hu) for evaluating our results on IRC-MM15-test. The authors also thank the anonymous reviewers for their insightful comments.


\bibliographystyle{IEEEtran}
\bibliography{IEEEabrv,research}

\begin{IEEEbiography}[{\includegraphics[width=1in,height=1.25in,clip,keepaspectratio]{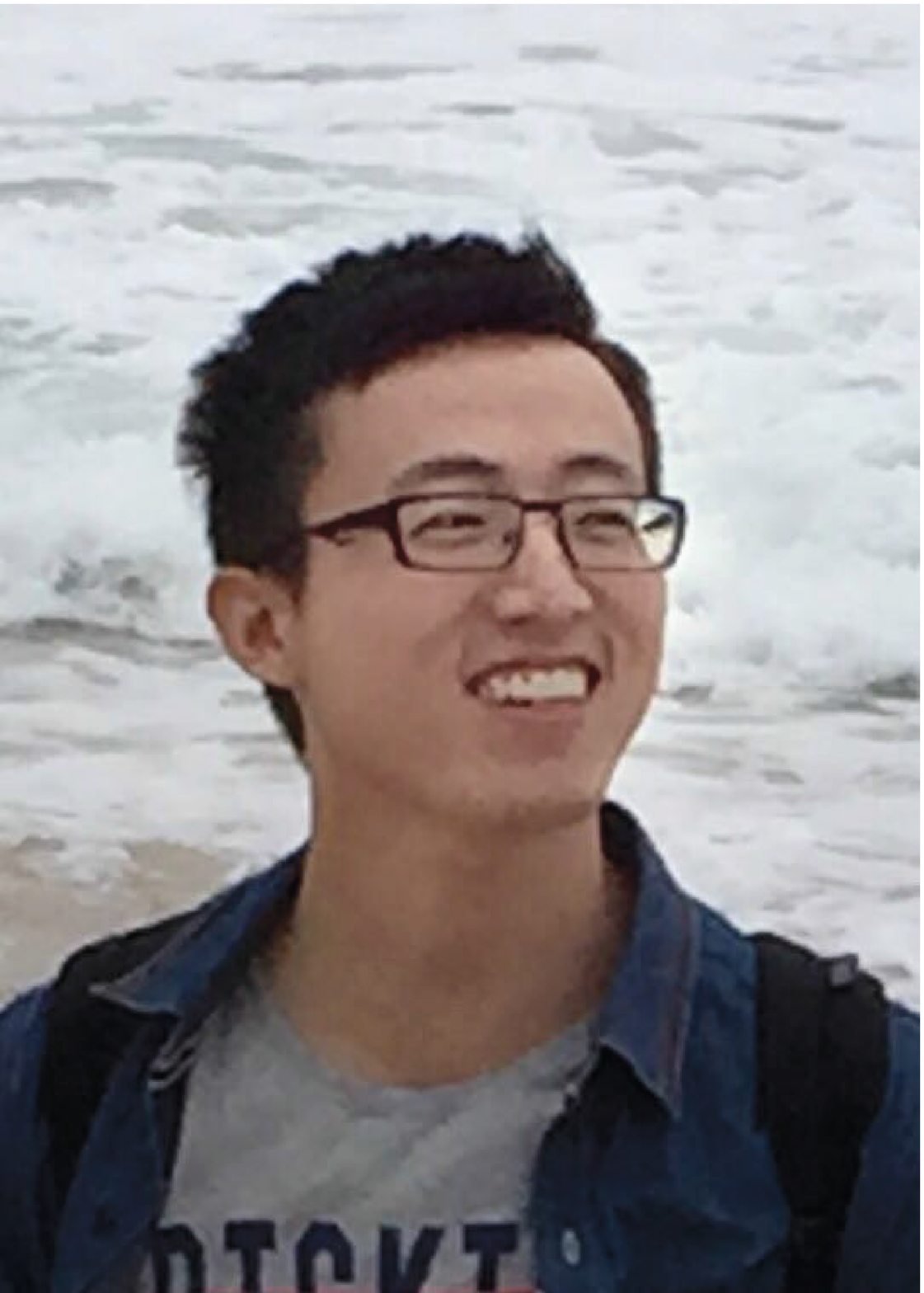}}]
{Jianfeng Dong} received the B.E. degree in software engineering from Zhejiang University of Technology, Hangzhou, China, in 2013.
He is currently a Ph.D. candidata in the School of Computer Science and Technology, Zhejiang University, Hangzhou, China.

His research interest is cross-media retrieval and deep learning. 
He was awarded the ACM Multimedia Grand Challenge Award in 2016.
\end{IEEEbiography}

\begin{IEEEbiography}[{\includegraphics[width=1in,height=1.25in,clip,keepaspectratio]{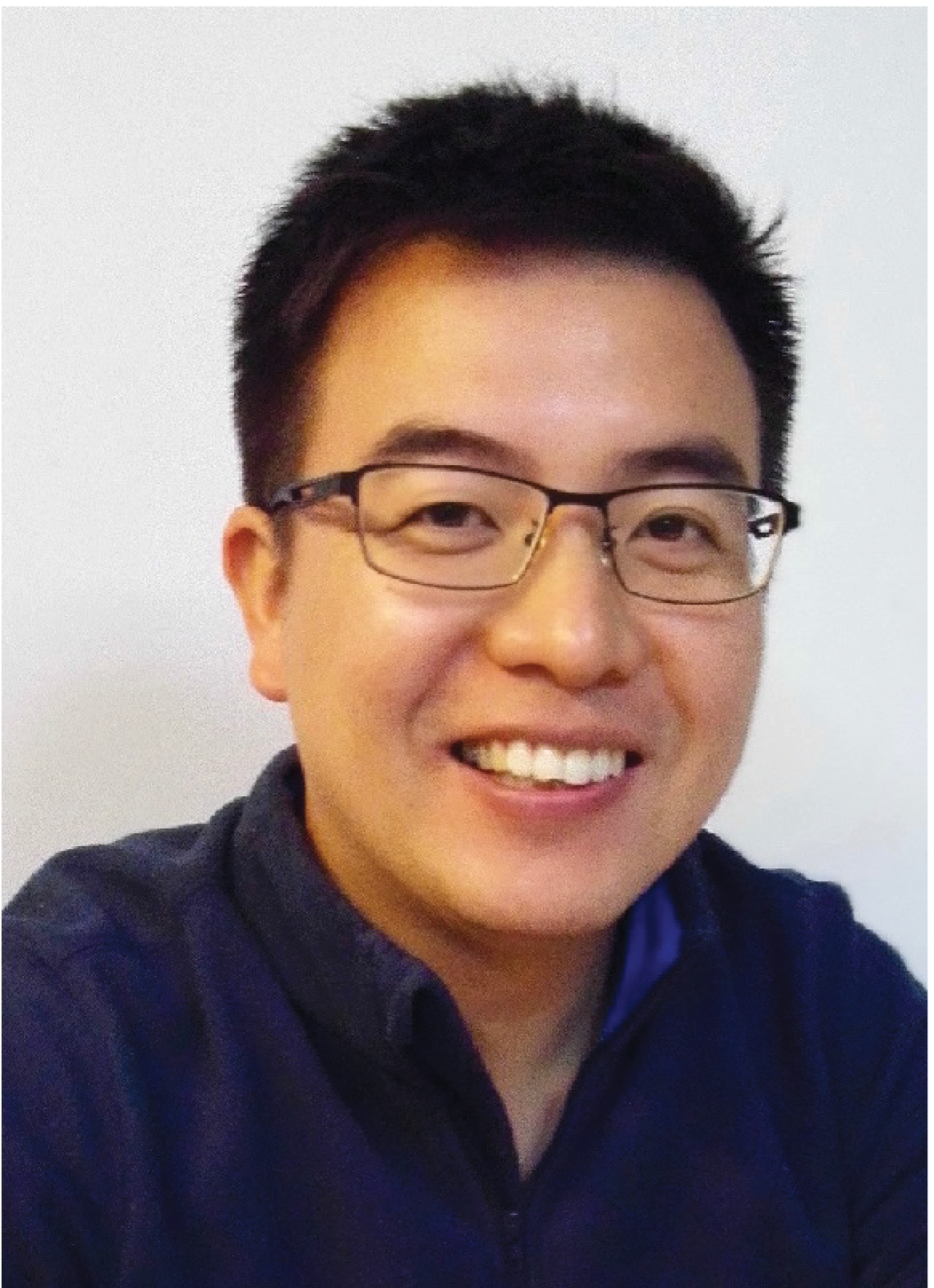}}]
{Xirong Li} received the B.S. and M.E. degrees from Tsinghua University, Beijing, China, in 2005 and 2007, respectively, and the Ph.D. degree from the University of Amsterdam, Amsterdam, The Netherlands, in 2012, all in computer science.

He is currently an Associate Professor with the Key Lab of Data Engineering and Knowledge Engineering, Renmin University of China, Beijing, China. His research includes image and video retrieval.

Prof. Li was an Area Chair of ICPR 2016 and Publication Co-Chair of ICMR 2015. He was the recipient of the ACM Multimedia 2016 Grand Challenge Award, the ACM SIGMM Best Ph.D. Thesis Award 2013, the IEEE TRANSACTIONS ON MULTIMEDIA Prize Paper Award 2012, the Best Paper Award of the ACM CIVR 2010, the Best Paper Runner-Up of PCM 2016 and PCM 2014 Outstanding Reviewer Award.

\end{IEEEbiography}


\begin{IEEEbiography}[{\includegraphics[width=1in,height=1.25in,clip,keepaspectratio]{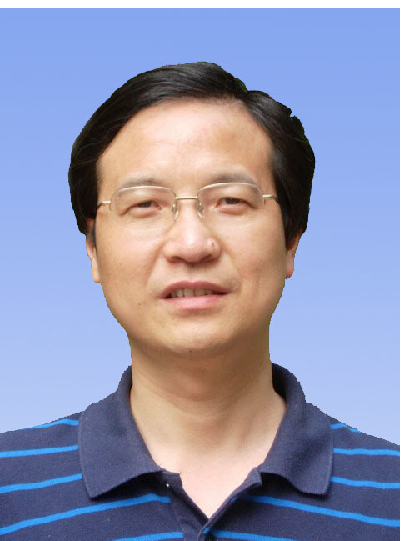}}]
{Duanqing Xu} received his Ph.D. degree in computer science from Zhejiang University, Hangzhou, China, in 2003. 

He is currently a professor at College of Computer Science, Zhejiang University. 
His main research interests include computer graphics, image processing, virtual reality and their applications. 

\end{IEEEbiography}


\vfill


\end{document}